\documentclass{article}

\usepackage{jmlr2e}
\usepackage{wrapfig}
\usepackage[utf8]{inputenc} 
\usepackage[T1]{fontenc}    
\usepackage{hyperref}       
\usepackage{url}  
\usepackage{diagbox}
\usepackage{wrapfig}
\usepackage{subfig}
\usepackage{booktabs}       
\usepackage{amsfonts}       
\usepackage{nicefrac}       
\usepackage{microtype}      

\newcommand{\argmin}{\arg\!\min}
\newcommand{\cnn}{f_{\mathrm{cnn}}}
\newcommand{\gru}{f_{\mathrm{gru}}}

\newcommand{\N}{\mathcal{N}}
\newcommand{\R}{\mathbb{R}}
\newcommand{\E}{\mathbb{E}}
\newcommand{\CNN}{CNN}
\newcommand{\PrCNN}{PrCNN}

\newcommand{\ACNN}{ACNN}
\newcommand{\PrACNN}{PrACNN}

\newcommand{\wendy}{\textcolor{red}{Wendy: }\textcolor{red}}

\usepackage[font={footnotesize}]{caption}
\usepackage{bbm}
\usepackage{hyperref}
\usepackage[utf8]{inputenc} 
\usepackage[T1]{fontenc}    
\usepackage{hyperref}       
\usepackage{url}            
\usepackage{booktabs}       
\usepackage{amsfonts}       
\usepackage{nicefrac}       
\usepackage{microtype}      
\usepackage{times}
\usepackage{graphicx} 
\usepackage{enumerate}
\usepackage{lipsum,multicol}
\usepackage{multirow}
\usepackage[bottom]{footmisc}
\usepackage{enumitem}
\usepackage{hhline}
\usepackage{xfrac}
\usepackage{array}
\usepackage{bbold}
\usepackage{float}
\usepackage{natbib}
\usepackage{amsthm}
\usepackage{amsfonts}
\usepackage{amsmath}
\usepackage{amssymb}
\usepackage{footnote}
\usepackage{braket}
\usepackage{algorithm2e}

\makesavenoteenv{tabular}
\makesavenoteenv{table}
\usepackage{color}
\usepackage{caption}

\usepackage[utf8]{inputenc} 
\usepackage[T1]{fontenc}    
\usepackage{hyperref}       
\usepackage{url}            
\usepackage{booktabs}       
\usepackage{amsfonts}       
\usepackage{nicefrac}       
\usepackage{microtype}      
\usepackage{makecell}

\title{Agent-Centric Representations for\\Multi-Agent Reinforcement Learning}

\author{%
  Wenling Shang\thanks{See project page \url{https://sites.google.com/view/agent-centric-marl} for the Supplementary Materials and videos. Correspondence: \texttt{wendyshang@google.com}. } \\
  DeepMind\\
  \\
  Lasse Espeholt, Anton Raichuk, Tim Salimans\\
  Google Research \\
}

\begin{document}

\maketitle

\begin{abstract}
Object-centric representations have recently enabled significant progress in tackling relational reasoning tasks. By building a strong object-centric inductive bias into neural architectures, recent efforts have improved generalization and data efficiency of machine learning algorithms for these problems. One problem class involving relational reasoning that still remains under-explored is multi-agent reinforcement learning (MARL). Here we investigate whether object-centric representations are also beneficial in the fully cooperative MARL setting. Specifically, we study two ways of incorporating an agent-centric inductive bias into our RL algorithm: 1. Introducing an agent-centric attention module with explicit connections across agents 2. Adding an agent-centric unsupervised predictive objective (i.e. not using action labels), to be used as an auxiliary loss for MARL, or as the basis of a pre-training step. We evaluate these approaches on the Google Research Football environment as well as DeepMind Lab 2D. Empirically, agent-centric representation learning leads to the emergence of more complex cooperation strategies between agents as well as enhanced sample efficiency and generalization. 
\end{abstract}
\section{Introduction}
Human perception and understanding of the world is structured around objects. Inspired by this cognitive foundation, many recent efforts have successfully built strong object-centric inductive biases into neural architectures and algorithms to tackle relational reasoning tasks, from robotics manipulation~\citep{devin2018deep}, to visual question answering~\citep{shi2019explainable}.  
%
Yet one problem class involving relational reasoning that still remains under-explored is multi-agent reinforcement learning (MARL). 

This work studies how agent-centric representations can benefit model-free MARL where each agent generates its policy independently. We consider a fully cooperative scenario, which can be modeled as a Multi-Agent Markov Decision Process (MAMDP), an extension of the single-agent MDP~\citep{boutilier1996planning}. 
%
In light of recent advances in model-free RL and neural relational reasoning~\citep{jaderberg2016reinforcement,zambaldi2018relational}, we study two ways of incorporating agent-centric inductive biases into our algorithm. 

First, we introduce an attention module~\citep{vaswani2017attention} with explicit connections across the decentralized agents.  
Existing RL works~\citep{zambaldi2018relational, mott2019towards, liu2019pic} have adapted similar self-attention modules on top of a single network. 
%
%
%
In our setup, the agents share a model to generate their actions individually to ensure scalability when the number of agents increases.
The proposed attention module is then implemented \emph{across} intermediate features from forward passes of the agents, explicitly connecting them. 
As we will show in experiments, this leads to the emergence of more complex cooperation strategies as well as better generalization.
The closest approach to ours is Multi-Attention-Actor-Critic (MAAC)~\citep{iqbal2018actor}, which also applies a shared encoder across agents and aggregates features for an attention module. 
However, each agent has its own unique critic that takes actions and observations of all agents through the attention features. 

Secondly, we develop an unsupervised trajectory predictive task\textemdash i.e. without using action labels\textemdash for pre-training and/or as an auxiliary task in RL.
Observations, without action labels, are often readily available which is a desirable property for pre-training.
Unlike prior works in multi-agent trajectory forecasting~\citep{yeh2019diverse, sun2019stochastic}, we consider an \emph{agent-centric} version where the location of each agent position is predicted separately.
This task encourages the model to reason over an agent's internal states such as its velocity, intent, etc. 
We explore two ways to leverage the pre-trained models in RL: (1) as weight initialization and (2) as a frozen column inspired by Progressive Neural Networks~\citep{rusu2016progressive}.
Furthermore, we investigate whether the agent-centric predictive objective serves as a suitable auxiliary loss for MARL. 
Auxiliary tasks have been used to facilitate RL representation learning in terms of stability and efficiency~\citep{oord2018representation, jaderberg2016reinforcement}. 

Our key contributions are as follows:
\begin{enumerate}
    \item  We introduce an agent-centric attention module for MARL to encourage complex cooperation strategies and generalization. We are the first to incorporate such attention module into an on-policy MARL algorithm.
    \item We employ an agent-centric predictive task as an auxiliary loss for MARL and/or for pre-training to improve sample efficiency. To our knowledge, we are the first to study auxiliary task in the context of MARL.
    \item We assess incorporating agent-centric inductive biases on MARL using the proposed approaches on challenging tasks from Google Research Football and DeepMind Lab 2D. 
    
\end{enumerate}
\begin{figure*}[t]
\centering
\subfloat[Baseline {\CNN}]{\label{fig:baseline}{\includegraphics[height=4.5cm]{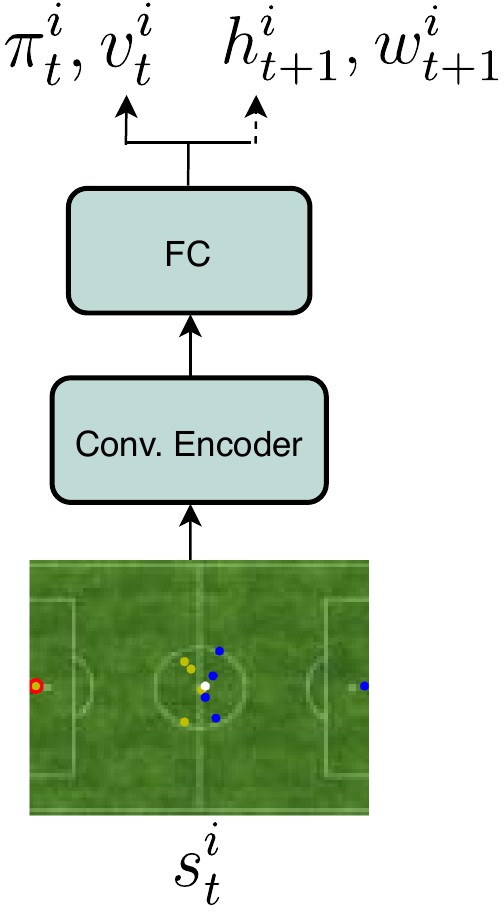}}}
\subfloat[{\ACNN} and agent-centric attention]{\label{fig:attention}{\includegraphics[height=4.5cm]{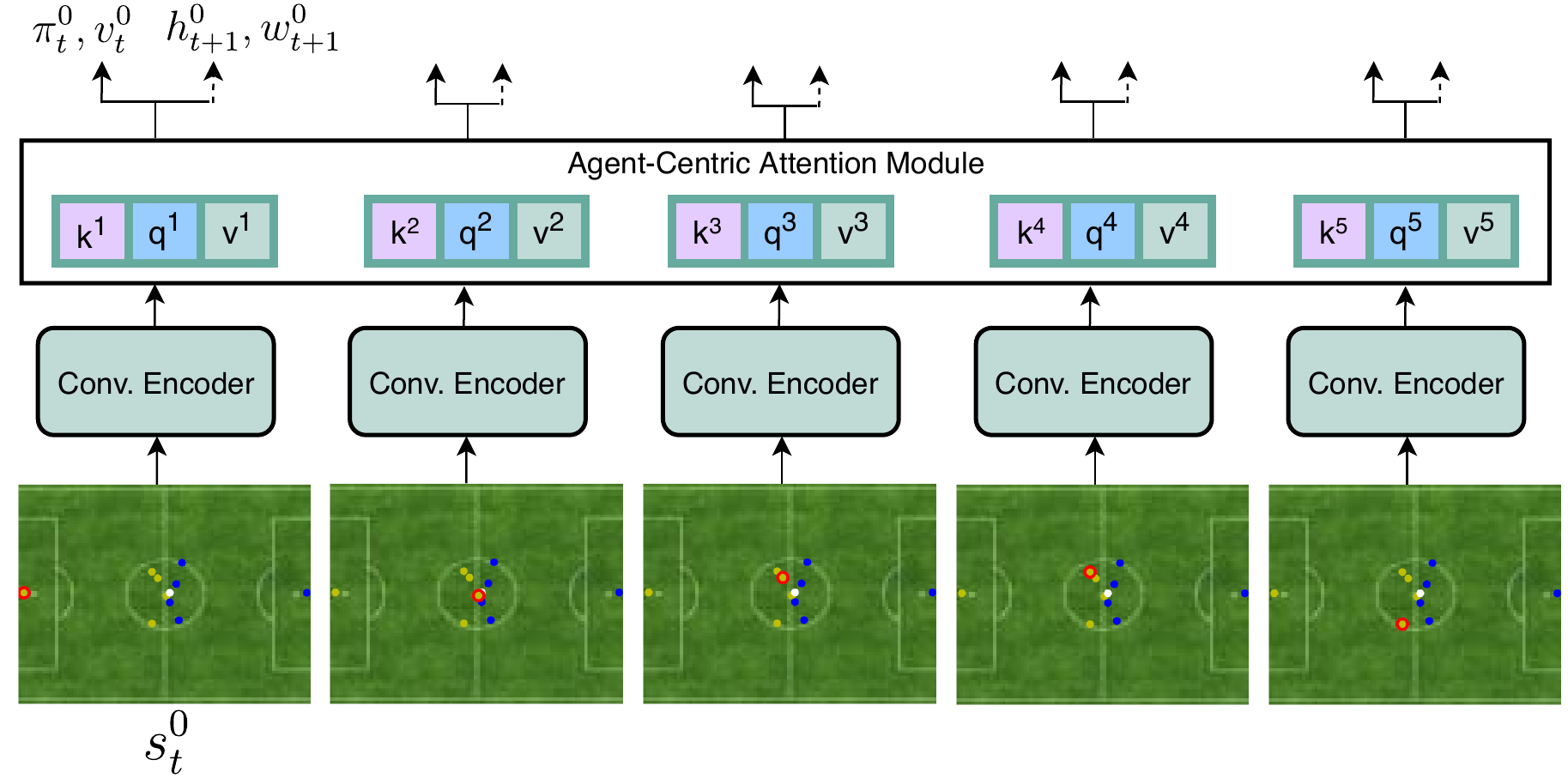}}}
\subfloat[{\PrCNN}]{\label{fig:progressive}{\includegraphics[height=4.2cm]{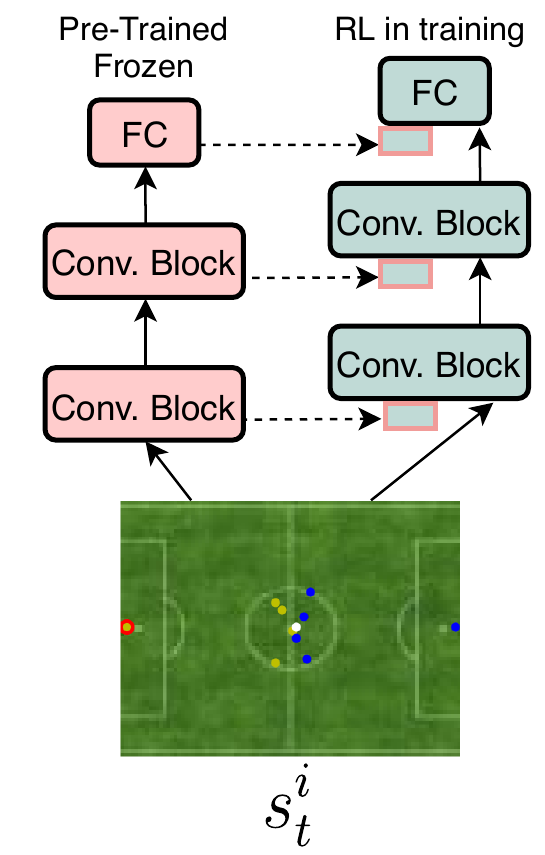}}}
\caption{(a) MARL {\CNN} baseline, agents share the model but with separate forward passes (b) {\ACNN} with the agent-centric attention module (c) Progressive {\CNN} ({\PrCNN}) with pre-trained model as a frozen column.}
\label{fig:model}
\end{figure*}
\section{Methods}
Section~\ref{sec:MARL} describes our problem setup, fully cooperative multi-agent reinforcement learning (MARL), in a policy gradient setting. Section~\ref{sec:attention} introduces the agent-centric attention module. Section~\ref{sec:observation} motivates the agent-centric prediction objective to learn from observations as an unsupervised pre-training step and/or as an auxiliary loss for MARL. 
%
\begin{figure*}[]
\centering
\includegraphics[width=0.99\textwidth]{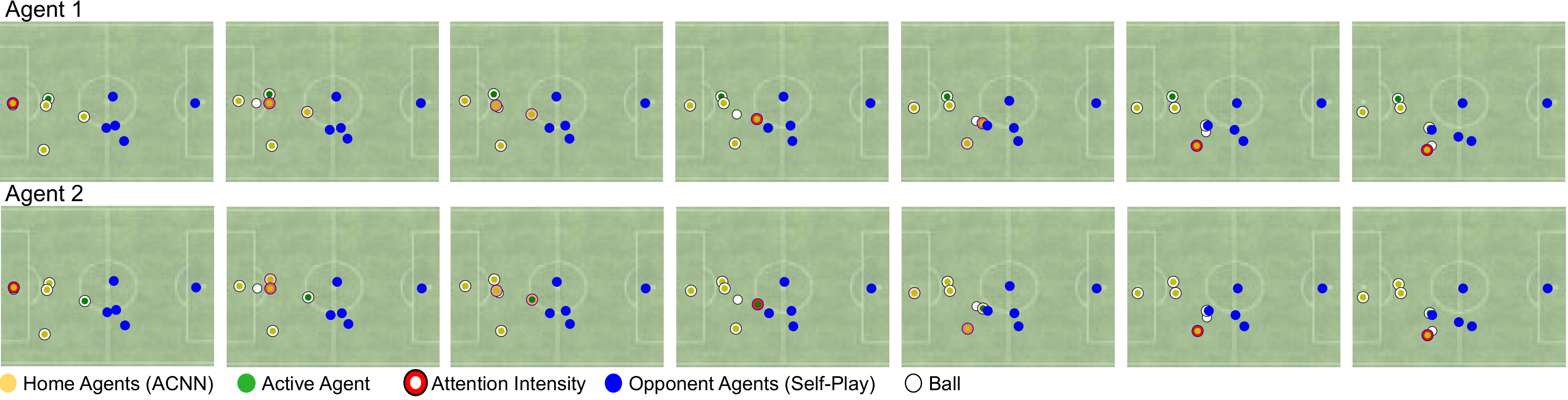}
\caption{Visualization of attention for two different home player agents. Green is the active agent being controlled, yellow are the other home agents, the red borders signify attention intensity, blue the opponents and white the ball. Attention is mostly on the agents around the ball but the attention weights of the two agents differ. E.g. in the 5th frame, one looks at agent possessing the ball, the other at the agent to whom it is passing the ball. Complete video replays of all players are on the project page.}
\label{fig:visualattention}
\end{figure*}
\subsection{Multiagent Reinforcement Learning}\label{sec:MARL}
%
We consider a group of $N$ agents, denoted by $\N$,  operating cooperatively in a shared environment towards a common goal. It can be formalized as a multi-agent Markov decision process (MAMDP)~\citep{boutilier1996planning}. A MAMDP is a tuple $(S, \{A^i\}_{i\in\N}, P, \{R^i\}_{i\in\N})$ where $S$ is the shared state space, $A^i$ the set of actions for the $i$-th agent, $\mathbf{A} = A^1 \times \cdots A^N$, $P: S\times \mathbf{A} \times S \to [0 , 1]$ the transition function, and $R: S \times \mathbf{A} \to \R $ the reward function. In our experiments, $S$ are 2D maps of the environments. 

We adapt an actor-critic method, V-trace~\citep{espeholt2018impala} implemented with SEED RL~\citep{espeholt2019seed}. The actor and critic are parameterized by deep neural networks. Each agent receives a state input $s^i$, covering the global $s$ and a specification of the agent's location, and generates its policy $\pi^i = \pi(s^i)$ and state value $V^i=V(s^i)$. 
The model is shared between agents, adding scalability when the number of agents grows larger and the environment more complex~\citep{iqbal2018actor, jiang2018learning, jeon2020scalable}. 
It also potentially alleviates instability due to the non-stationary nature of multi-agent environments by sharing the same embedding space~\citep{lowe2017multi}. 

The goal for all agents is to maximize the expected long term discounted global return 
\begin{equation}
J(\theta)=\E_{s\sim d, \mathbf{a} \sim \pi}\left[ \Sigma_{t\geq 0} \gamma^t R(s_t,\mathbf{a})\right] = \E_{s\sim d} \left[ V(s)\right] ,
\end{equation}
where $0\leq \gamma \leq 1$ is the discount factor and $\mathbf{a} {=} (a^1,{\cdots}, a^N)$,
\begin{equation}
V(s) = \E_\pi\left[\Sigma_{t\geq t_0} \gamma^t R(s_t,\mathbf{a}_t)| s_0 = s\right] 
\end{equation}
the state value, and $\theta$ the parameterization for policy and value functions. 
In a decentralized cooperative setting, ~\cite{zhang2018fully} proves the policy gradient theorem
\begin{equation}
    \nabla_\theta J(\theta) {=} \E_{d,\pi} \left[\nabla \log \pi^i(s, a^i) Q(s, \mathbf{a})  \right] {=} \E_{d, \pi} \left[\nabla \log \pi(s, \mathbf{a}) A(s, \mathbf{a})  \right],\nonumber
\end{equation}
where $Q(s,\mathbf{a}) {=} \E_\pi \left[R(s_t,\mathbf{a}_t) {+} \gamma V(s_{t+1})\right]$ is the state-action value and $A(s,\mathbf{a}){=}Q(s,\mathbf{a}){-}V(s)$ the advantage for variance reduction. 
In theory, $V^i {\approx} V$ and we can directly apply V-trace for each individual agent with the policy gradient above.
In practice, we slightly shape each agent's reward (see Section~\ref{sec:env}) but the policy gradient direction is approximately retained (proof in Appendix 1).  
\subsection{Agent-Centric Attention Module}\label{sec:attention}
In the MARL baseline {\CNN} (Figure~\ref{fig:baseline}), 
an agent makes independent decisions without considering the high-level representations from other agents. 
%
This can hinder the formation of more complex cooperation strategies. 
To address the issue, we propose a novel attention module built upon the multi-head self-attention mechanism~\citep{vaswani2017attention} to explicitly enable relational reasoning across agents.
%
%
%
As shown in Figure~\ref{fig:attention} and Equation~\ref{eq:attn}, the forward pass of each agent produces the key, query and value independently, which are congregated to output the final features for each agent. These features are then sent to the RL value and policy heads.
The attention module allows easy model interpretation~\citep{mott2019towards}, see Figure~\ref{fig:visualattention} with a detailed explanation in Section~\ref{sec:mainexp}.
%
We term the model {\ACNN}. 

%
%
%
%
%
%
%
%

The following summarizes the operations of our attention module:
\begin{align}\label{eq:attn}
    &z^i {=} f_{\mathrm{fc}}(\cnn(s^i)), q^i {=} q\left(\mathrm{LN}(z^i)\right), k^i {=}k\left(\mathrm{LN}(z^i)\right), v^i {=} v\left(\mathrm{LN}(z^i)\right);\nonumber\\
    &K {=} (k^1 \cdots k^N), Q {=}(q^1\cdots q^N), V {=} (v^1 \cdots v^N), \nonumber \\
    &\widetilde{V} = \mathrm{Attn}(Q,K,V){=}\sigma(\frac{QK^T}{\sqrt{d_k}})V; \nonumber \\
    &\tilde{z}^i = \mathrm{LN}(z^i + \tilde{v}^i) \longrightarrow \pi^i = f_\pi(\tilde{z}^i), v^i = f_v(\tilde{z}^i).
\end{align}

We find a proper placement of Layer Normalization~\citep{ba2016layer} within the attention module is crucial.

\subsection{Multi-agent Observation Prediction}\label{sec:observation}
%
When developing a new skill, humans often extract knowledge by simply observing how others approach the problem. We do not necessarily pair the observations with well-defined granular actions but grasp the general patterns and logic.
%
%
Observations without action labels are often readily available, such as recordings of historical football matches and team-based FPS gameplay videos. 
Therefore, in this work we explore how to transfer knowledge from existing observations to downstream MARL without action labels being present. 

A useful supervision signal from observation is the agent's  location. Even when not directly accessible, existing techniques~\citep{he2017mask,ren2015faster} can in many cases extract this information. We thus adopt an agent's future location as a prediction target. It is worth noting that the same action can lead to different outcomes for different agents depending on their internal states such as velocity. Therefore, we expect the location predictive objective provides cues, independent of actions taken, for the model to comprehend an agent's intent and its internal states. 

Recent works~\cite{yeh2019diverse, sun2019stochastic, zhan2018generative} develop models that predict trajectories over all agents at once.
We, on the other hand, task each agent to predict its future location, arriving at an agent-centric predictive objective, as illustrated in Figure~\ref{fig:baseline}.
The motivation for the agent-centric objective is two-fold. 
For one, as discussed later in this section, the agent-centric loss can be integrated as an auxiliary loss to MARL in a straightforward manner. 
%
And secondly, if the model predicts for all agents at the same time, it can overwhelm the RL training as later compared in Section~\ref{sec:observeall}. 

Concretely, after collecting observation rollouts, we minimize a prediction loss over the observations instead of maximizing return as in RL training (for details see Section~\ref{sec:observationlearning}). Our observation is in a 2D map format, and the prediction task for each agent consists of predicting location heatmaps along height $\sigma_h^i$ and width $\sigma_w^i$ as softmax vectors. We train these predictions by minimizing the negative log likelihood via cross-entropy loss, 
\begin{equation}
\argmin_\theta \E_{\sigma ^i_h}\left[- \log \sigma^i_h[h^i_{t+1}]\right], \argmin_\theta \E_{\sigma ^i_w}\left[- \log \sigma^i_w[w^i_{t+1}]\right],
\end{equation}
where $h^i_{t+1}$ and $w^i_{t+1}$ are the ground truth next step locations of the $i$th agent. 
The pre-training is applied to both the {\CNN} and {\ACNN} architectures. 
Next, we investigate two ways to transfer knowledge from the pre-trained models to MARL. 

\textbf{Weight Initialization} Transfer via weight initialization is a long-standing method for transfer learning in many domains~\citep{donahue2013decaf, devlin2018bert}. Often, the initialization model is trained on larger related supervised task~\citep{donahue2013decaf,carreira2017quo}, but unsupervised pre-training~\citep{he2019momentum, devlin2018bert} has also made much progress. In RL, some prior works initialize models with weights trained for another relevant RL task, but general pre-training through a non-RL objective has been explored less. 
%
%
%

\textbf{Progressive Neural Networks} Progressive Neural Networks~\citep{rusu2016progressive} are originally designed to prevent catastrophic forgetting and to enable transfer between RL tasks. They build lateral connections between features from an existing RL model\textemdash on a relevant task\textemdash to the RL model in training. The setup is also a promising candidate for transferring knowledge from our pre-trained models to MARL. Specifically, a pre-trained model becomes a \emph{frozen column} from which the intermediate features, after a non-linear function, are concatenated to the corresponding activation from the RL model, as shown in Figure~\ref{fig:progressive}. We experiment with Progressive Networks combined with {\CNN} and {\ACNN}, called {\PrCNN} and {\PrACNN}.
\subsubsection{Multi-agent Observation Prediction as Auxiliary Task for MARL}\label{sec:aux}
Auxiliary objectives for single-agent RL in terms of stability, efficiency and generalization have been widely studied~\citep{oord2018representation, jaderberg2016reinforcement}.
In light of prior works, we assess using the agent-centric prediction objective as an auxiliary task for MARL. 
Thanks to the convenient formulation of the agent-centric objective, we can simply add prediction heads in juxtaposition with the policy and value heads as in Figure~\ref{fig:baseline}.
%

\section{Environments and Setups}\label{sec:env}
We consider two multi-agent domains, Google Research Football~\citep{kurach2019google} (GRF) and DeepMind Lab 2D~\citep{dmlab2d} (DMLab2D), with a focus on the more dynamically complex GRF. 
\subsection{Google Research Football}\label{sec:grf}
Google Research Football~\citep{kurach2019google} (GRF) is a FIFA-like environment. Home player agents match against the opponent agents in the game of 5-vs-5 football. All agents are individually controlled except for the goalie, which by default is operated under the built-in rules. Each agent picks 1 out of 19 actions to execute at each step, with a total of 3000 frames per game. We compete against two types of opponent policies: the built-in AI and the self-play AI. 

The built-in AI is a rule-based policy~\citep{gameplayfootball}\footnote{To add more challenge when training against built-in AI, we also take control over the goalie.} Its behavioral patterns reflect simple and logical football strategies. But the logic can be exploited easily. For instance, the video in the Supplementary Materials show that the built-in AI does not have programmed rules to prevent getting into an offside position. Therefore, the learned policy takes advantage of this flaw to switch possession by tricking the built-in AI into off-set positions instead of actual defending. 

The other more robust and generalized opponent policy is trained via self-play
(for more details in Table~\ref{tab:tournament} and Appendix 3). It requires more advanced cooperative strategies to win against the self-play AI. 

Additionally, in our ablation studies, we use another a single-agent setup, 11-vs-11 ``Hard Stochastic''~\citep{kurach2019google}. In this case, one active player is being controlled at one time. 

For all setups, the home agents are all rewarded ${+}1$ after scoring and ${-}1$ if a goal is conceded. To speed up training, we reward the agent in possession of the ball an additional ${+}0.1$ if it advances towards the opponent's goal~\citep{kurach2019google}.  

Observations are in the Super Mini Map (SMM)~\citep{kurach2019google} format. An SMM is a 2D bit map of spatial dimension $72\times 96$, covering the entire football field. It is composed of four channels: positions of the home players, the opponents, the ball and the active agent. SMMs across four time steps are stacked to convey information such as velocity. When learning from observations, we predict the height and width locations of an agent on the SMM by outputting $72$- and $96$-dim heatmap vectors separately. 
\subsection{DeepMind Lab 2D}
\vspace{-0.05in}
\begin{wrapfigure}{r}{0.3\textwidth}
  \vspace{-.7cm}
  \begin{center}
    \includegraphics[width=0.25\textwidth]{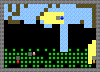}
  \end{center}
  \vspace{-0.1in}
  \caption{DMLab2D ``Cleanup''}\label{fig:cleanup}
\end{wrapfigure}
DeepMind Lab 2D (DMLab2D)~\citep{dmlab2d} is a framework supporting multi-agent 2D puzzles. We experiment on the task ``Cleanup'' (Figure~\ref{fig:cleanup}). At the top of the screen, mud is randomly spawned in the water and at the bottom apples spawned at a rate inversely proportional to the amount of mud. The agents are rewarded ${+}1$ for each apple eaten. Four agents must cooperate between cleaning up and eating apples. The list of actions is in Appendix 2. 
There are 1000 $72\times 96$ RGB frames per episode. The state input for each agent colors itself blue and the others red. Frames across four time steps are stacked for temporal information.

\subsection{Implementation Details}\label{sec:implement}
For all MARL experiments, we sweep the initial learning rate over the set of values $(0.00007, 0.00014, 0.00028)$ and sweep the auxiliary loss coefficient over $(0.0001, 0.0005)$. The loss coefficient for value approximation is $0.5$. We use TensorFlow~\cite{abadi2016tensorflow} default values for other parameters in ADAM~\cite{kingma2014adam}. The unroll length is $32$, discounting factor $0.99$, entropy coefficient in V-trace $0.0005$. The multi-agent experiments for GRF are run on 16 TPU cores, 10 of which are for training and 6 for inference, and 2400 CPU cores for actors with 18 actors per core. The batch size is 120. We train 500M frames when playing against the built-in AI and 4.5G frames when playing against the built-in AI. The single-agent experiments (Section~\ref{sec:single}) for GRF are run on 32 TPU cores, 20 of which for training and the rest for inference, and 480 CPU cores for actors with 18 actors per core. The batch size is 160. We train 900M frames for the single-agent RL tasks. The DMLab2D experiments are run on 8 TPU cores, 5 of which for training the other for inference, and 1472 CPU cores with 18 actors per core. 

\begin{figure*}[]
\centering
\includegraphics[width=0.24\textwidth]{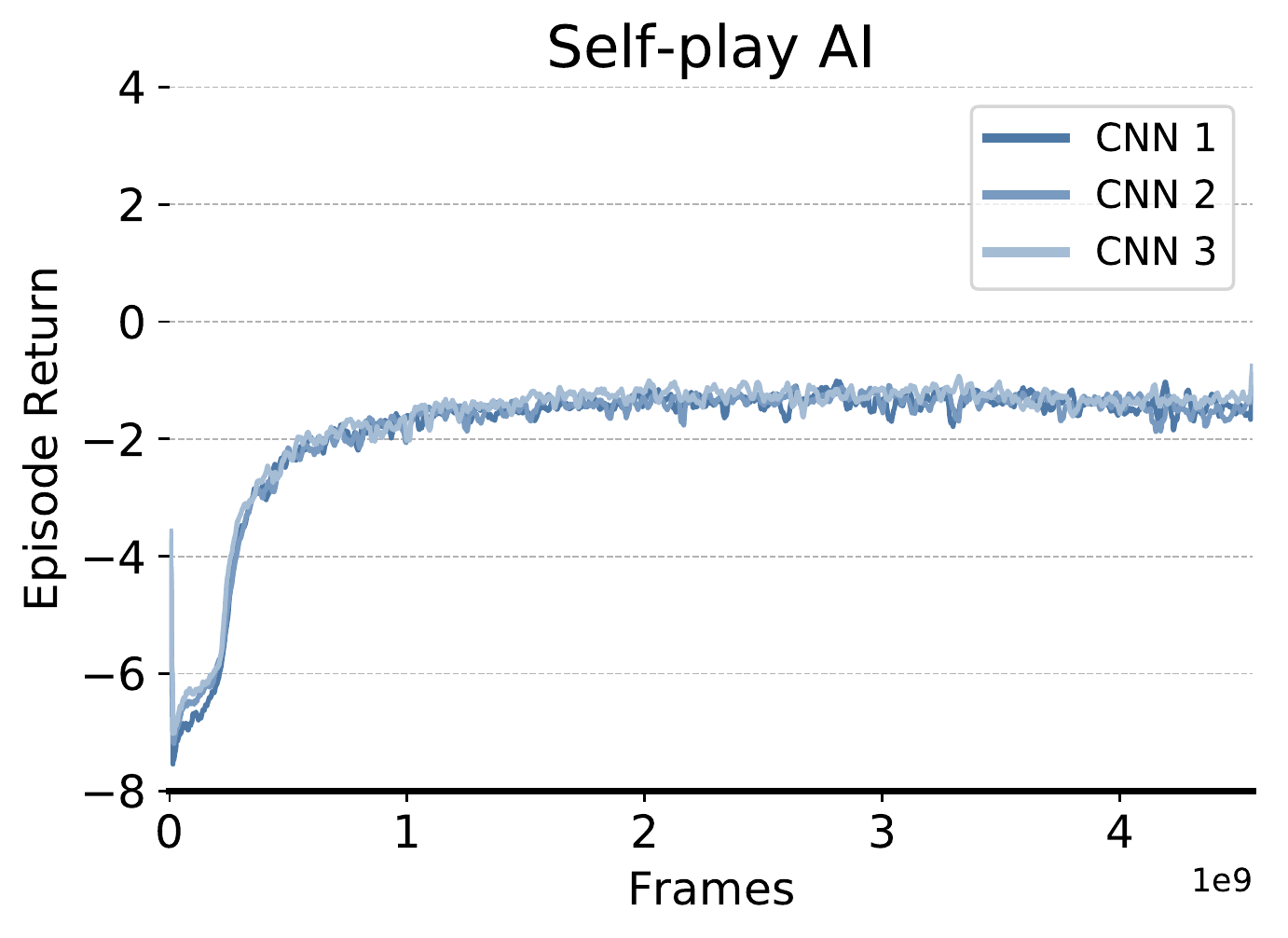}
\includegraphics[width=0.24\textwidth]{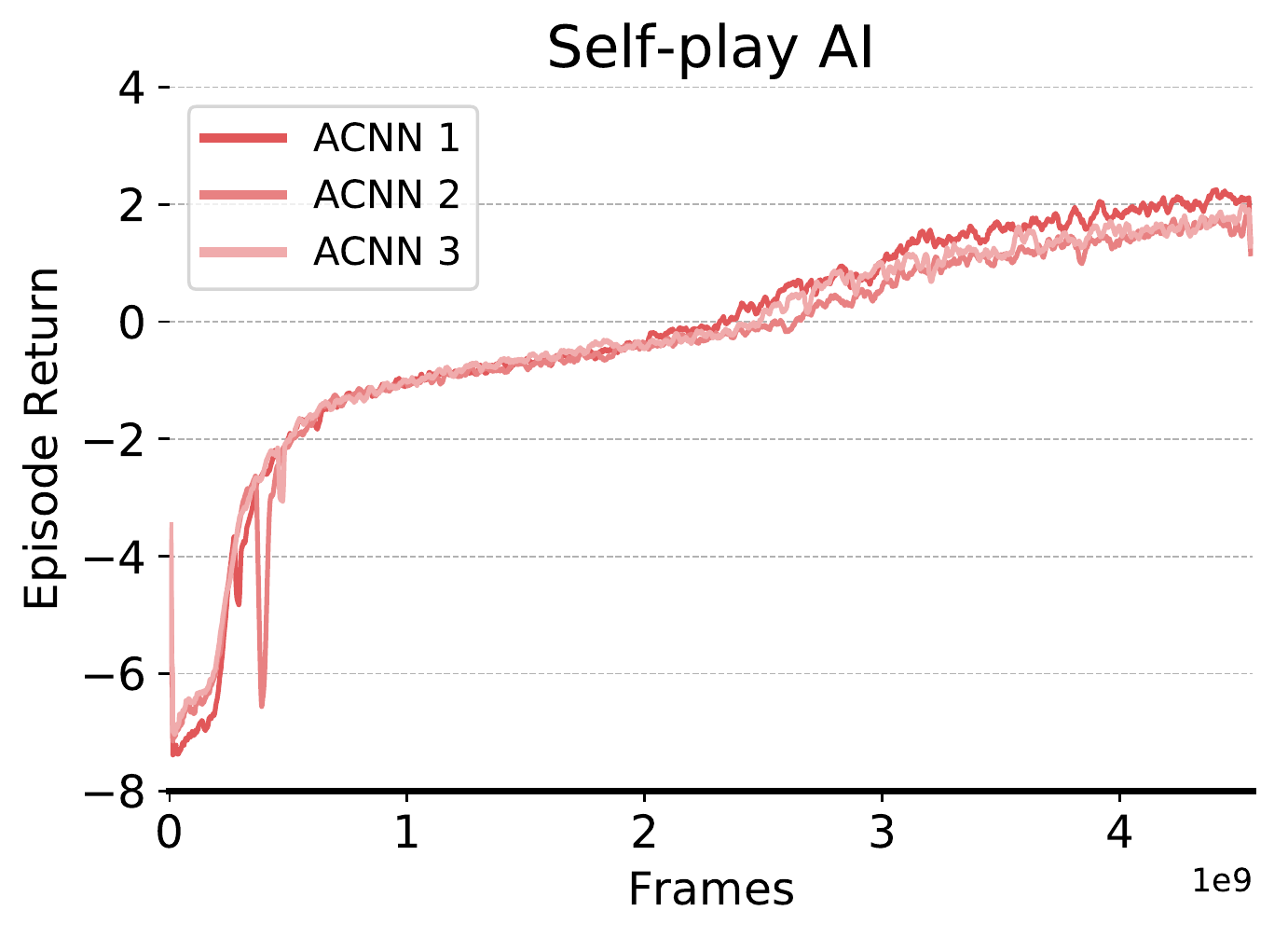}
\includegraphics[width=0.24\textwidth]{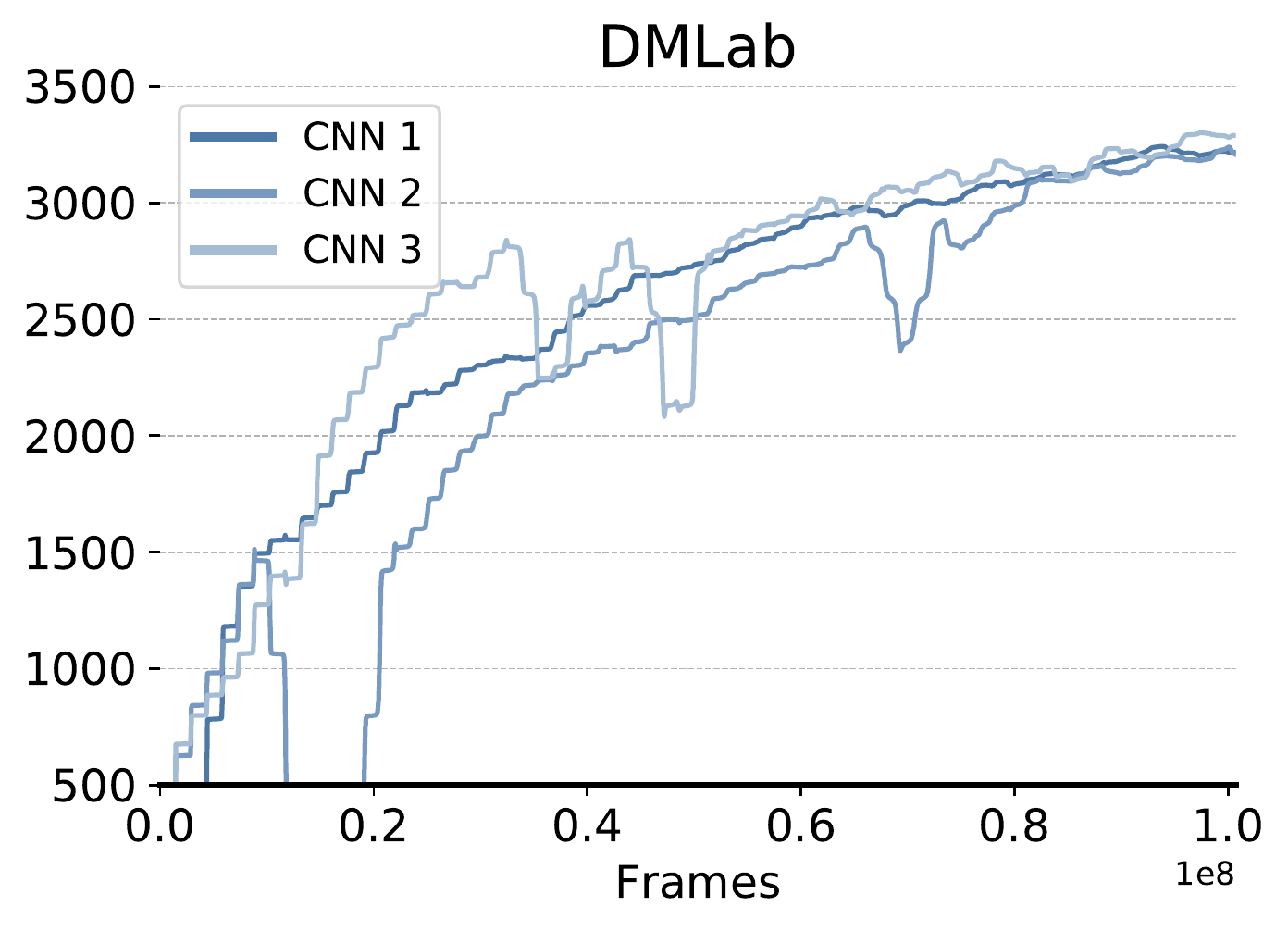}
\includegraphics[width=0.24\textwidth]{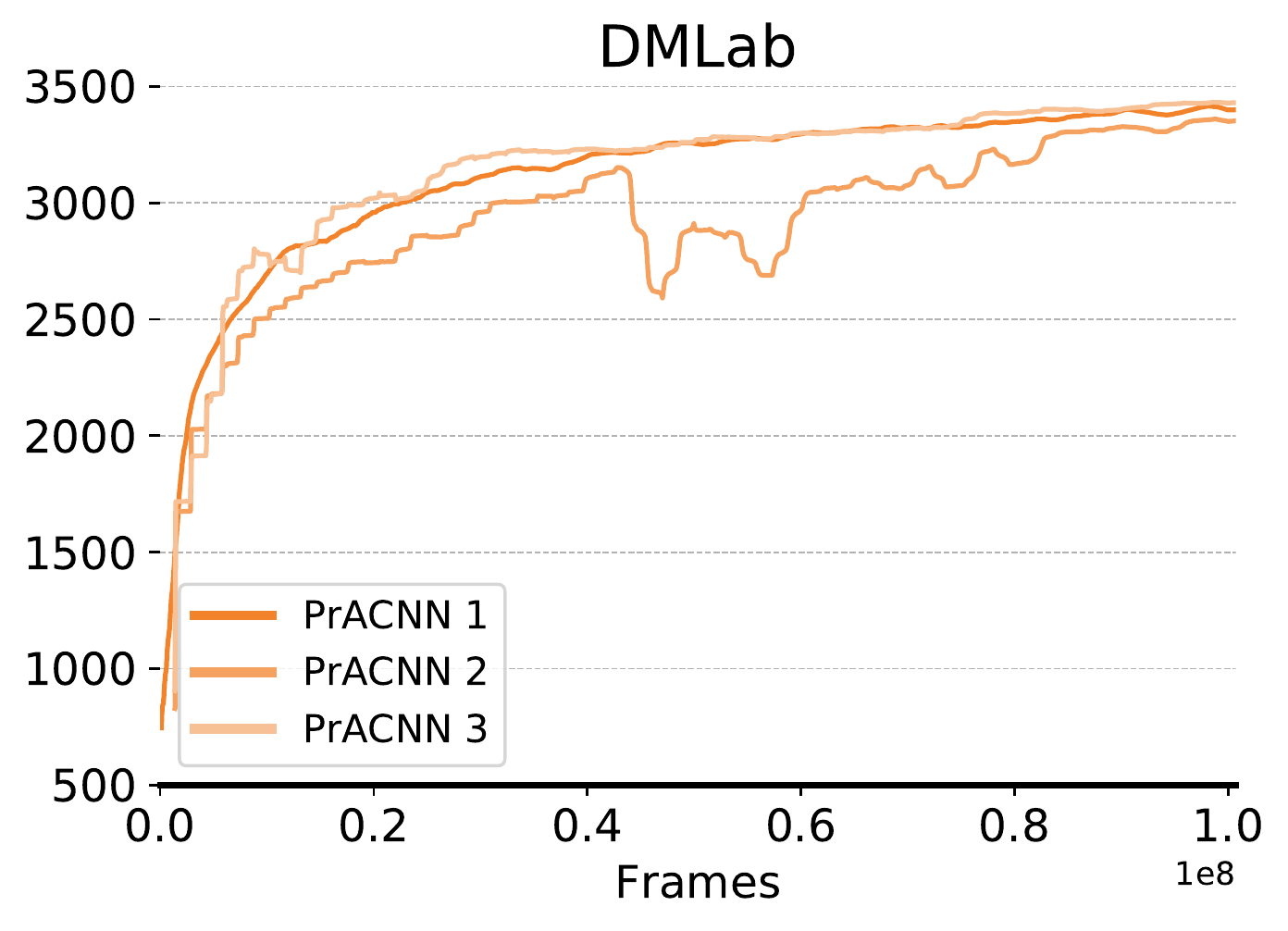}
\caption{We plot out all 3 runs for selected models on GRF self-play and DMLab2D to showchase that our environments are relatively stable to random seeds.}
\label{fig:seeds}
\vspace{-0.1in}
\end{figure*}

All learning curves presented in the paper are of 3 seeds. Our environments are relatively stable to random seeds. We display all 3 runs for selected models on GRF self-play and DMLab2D as examples in Figure~\ref{fig:seeds}. Therefore, we plot the average over 3 seeds and omit the standard error bars for clearer exposition. 
\begin{figure*}[]
\centering
\includegraphics[width=0.28\textwidth]{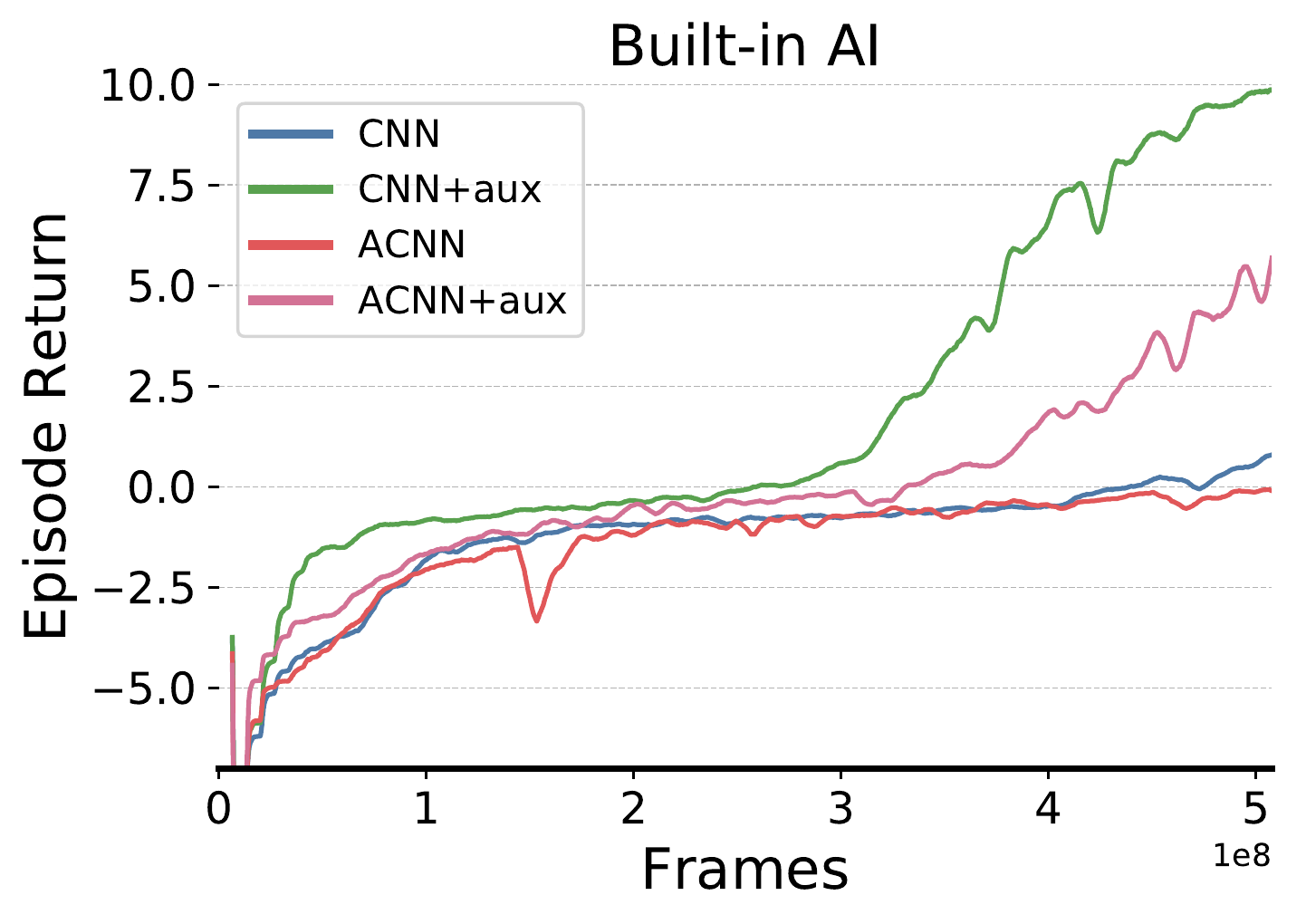}\hspace{0.35in}
\includegraphics[width=0.28\textwidth]{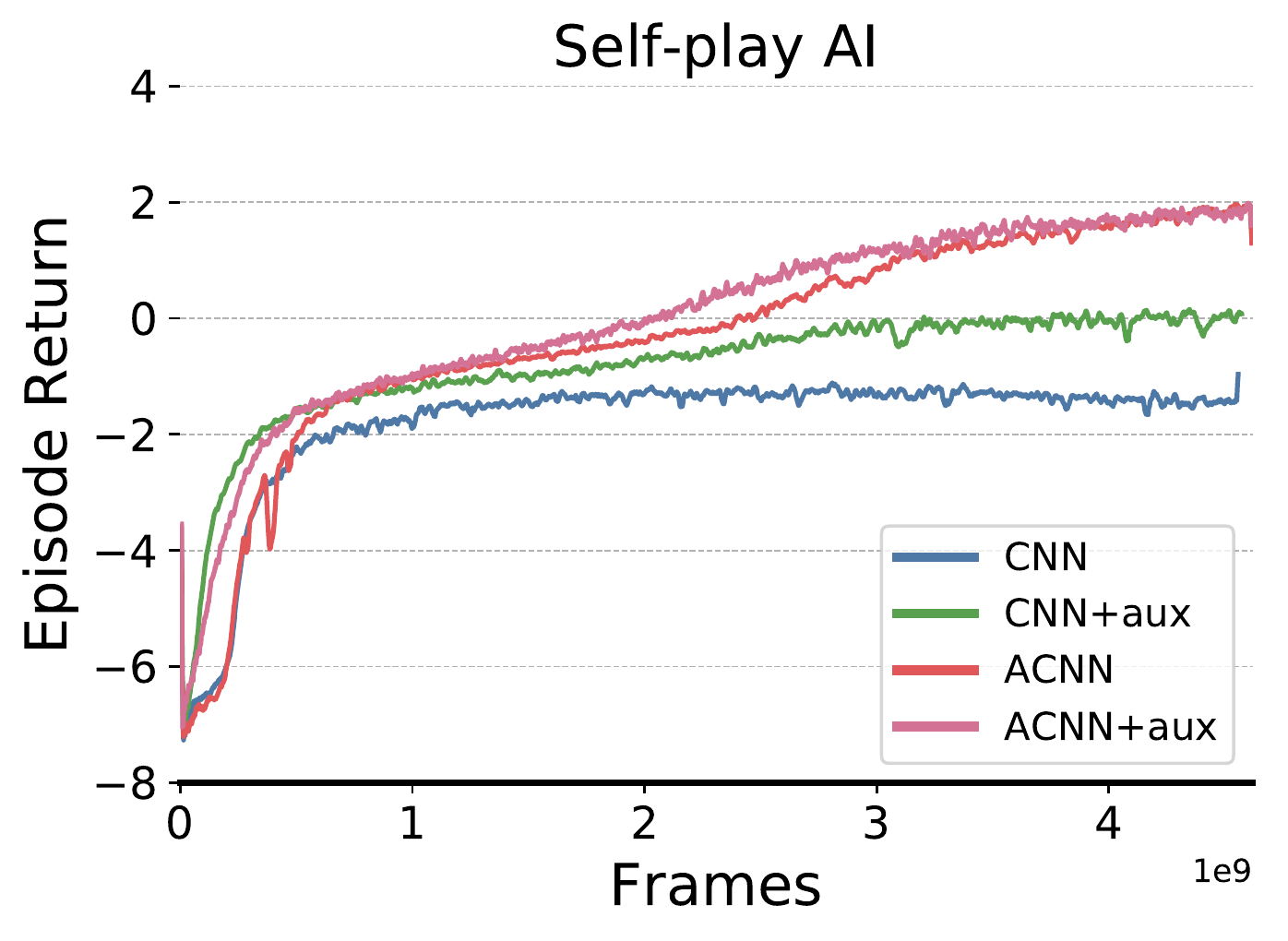}\hspace{0.35in}
\includegraphics[width=0.28\textwidth]{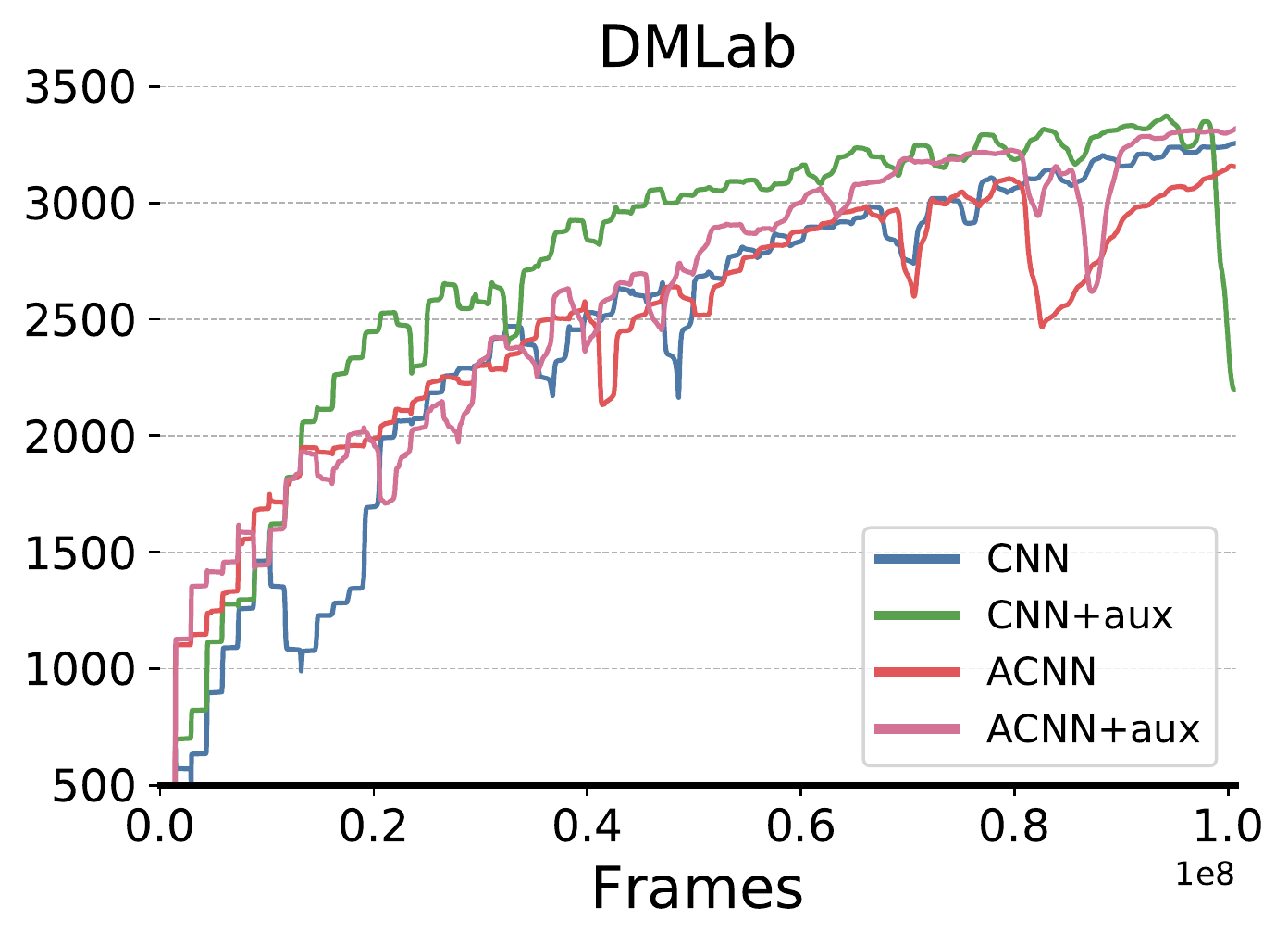}
\caption{Comparison of  {\CNN} and {\ACNN} models trained without and with the auxiliary loss. On the most challenging task self-play GRF which requires more advanced relational reasoning, attention becomes crucial\textemdash without it, the average return does not pass the zero mark. On both GRF tasks, the auxiliary loss improves performance and sample efficiency. Plots are averaged over 3 random seeds.}
\label{fig:aux}
\end{figure*}
\section{Experiments and Discussions}
This section starts with describing our model architectures, implementation details for pre-training from observations and implementation details for MARL. Next, we dive into analysing the efficacy of agent-centric representations for MARL. Finally, we conduct two additional ablation studies: comparing agent-centric representation learning against an agent-agnostic alternative on MARL, and evaluating agent-centric representation learning on single-agent RL. 
%
\subsection{Model Architecture} 
The baseline in our experiments is a convolutional neural network (CNN) followed by a fully connected (FC) layer (Figure~\ref{fig:baseline}). The policy, value and height/width predictive heads are immediately after the FC layer. In experiments with the attention module, the FC layer is replaced by a 2-head attention module as illustrated in Figure~\ref{fig:model} and Equation~\ref{eq:attn}. The detailed architectures are adapted from~\cite{espeholt2019seed} and attached in the Supplementary Materials.
\begin{figure*}[]
\centering
\subfloat[Comparison of {\CNN}, {\CNN} using initialization, and {\PrCNN}.]{\label{fig:init1}{
\includegraphics[width=0.28\textwidth]{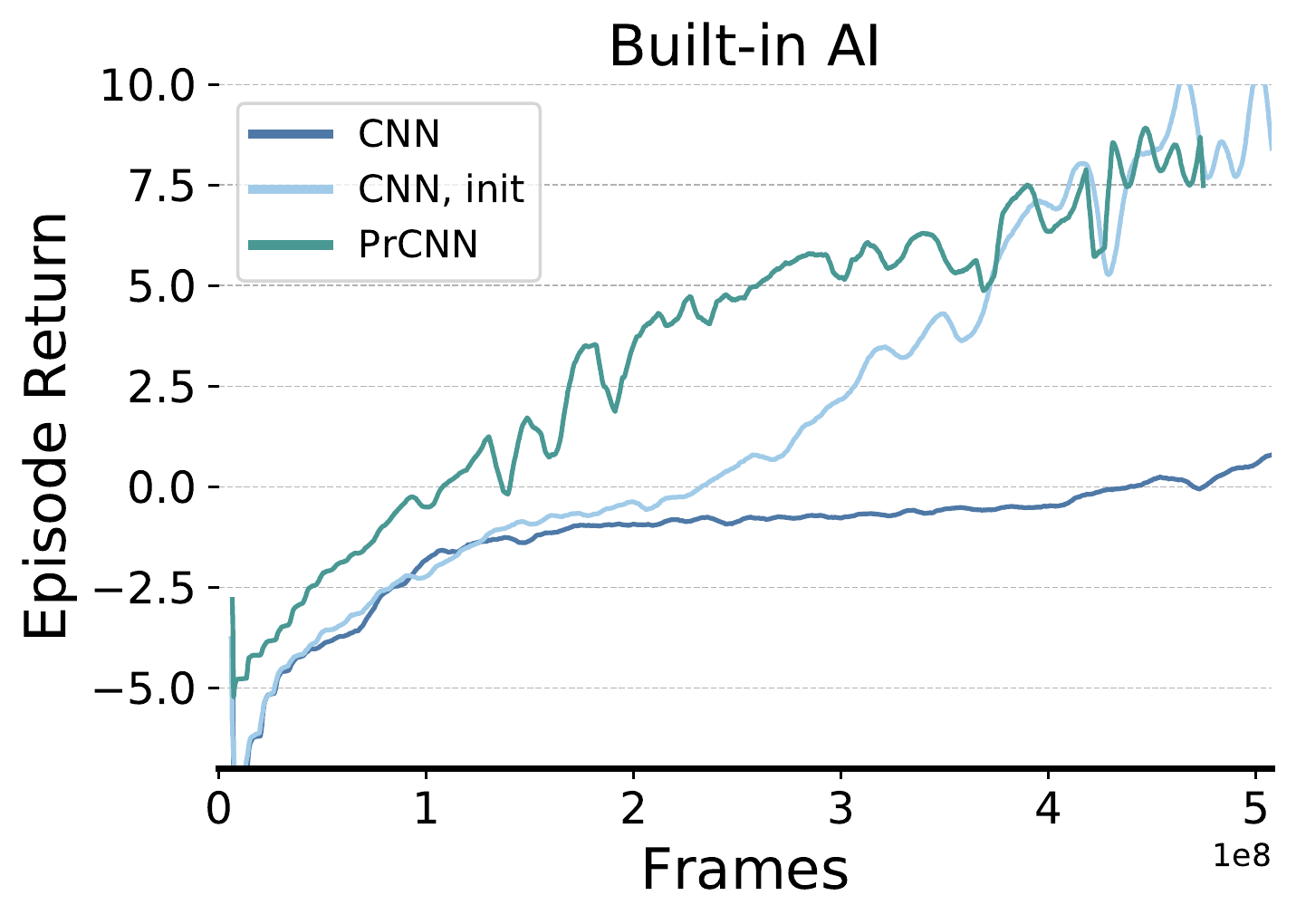}\hspace{0.35in}
\includegraphics[width=0.28\textwidth]{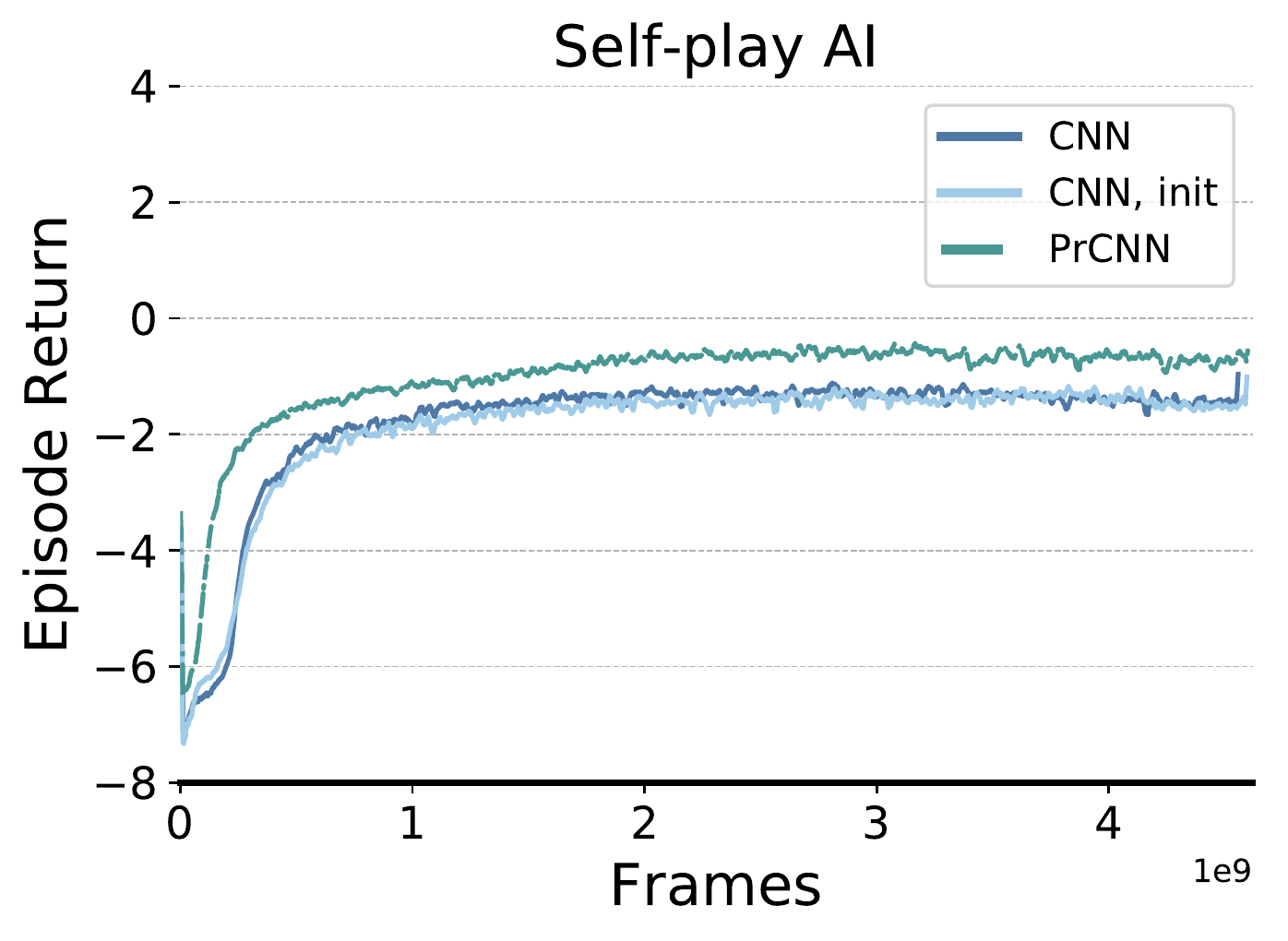}\hspace{0.35in}
\includegraphics[width=0.28\textwidth]{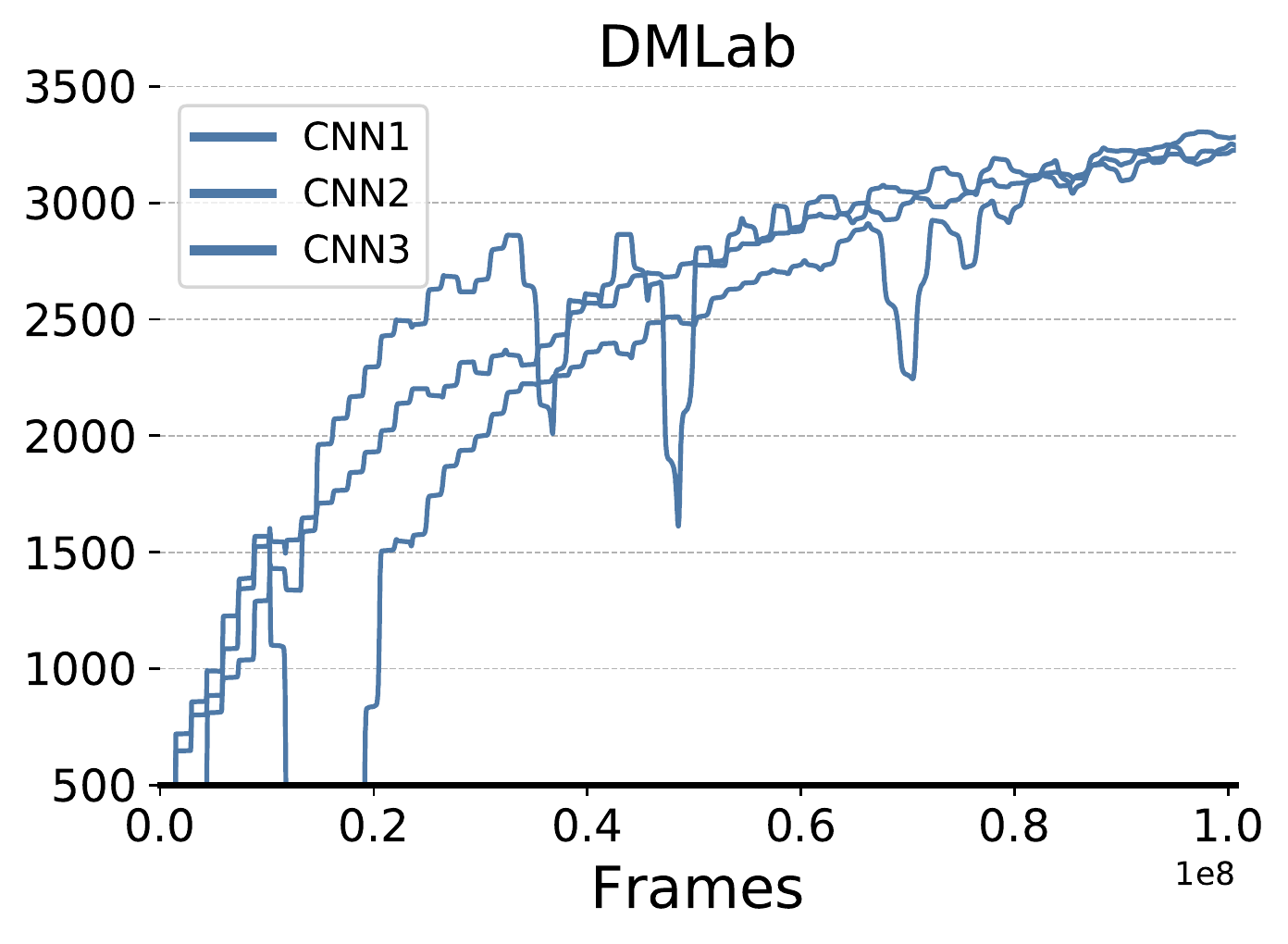}}}\\
\subfloat[Comparison of {\ACNN}, {\ACNN} using initialization, and {\PrACNN}.]{\label{fig:init2}{
\includegraphics[width=0.28\textwidth]{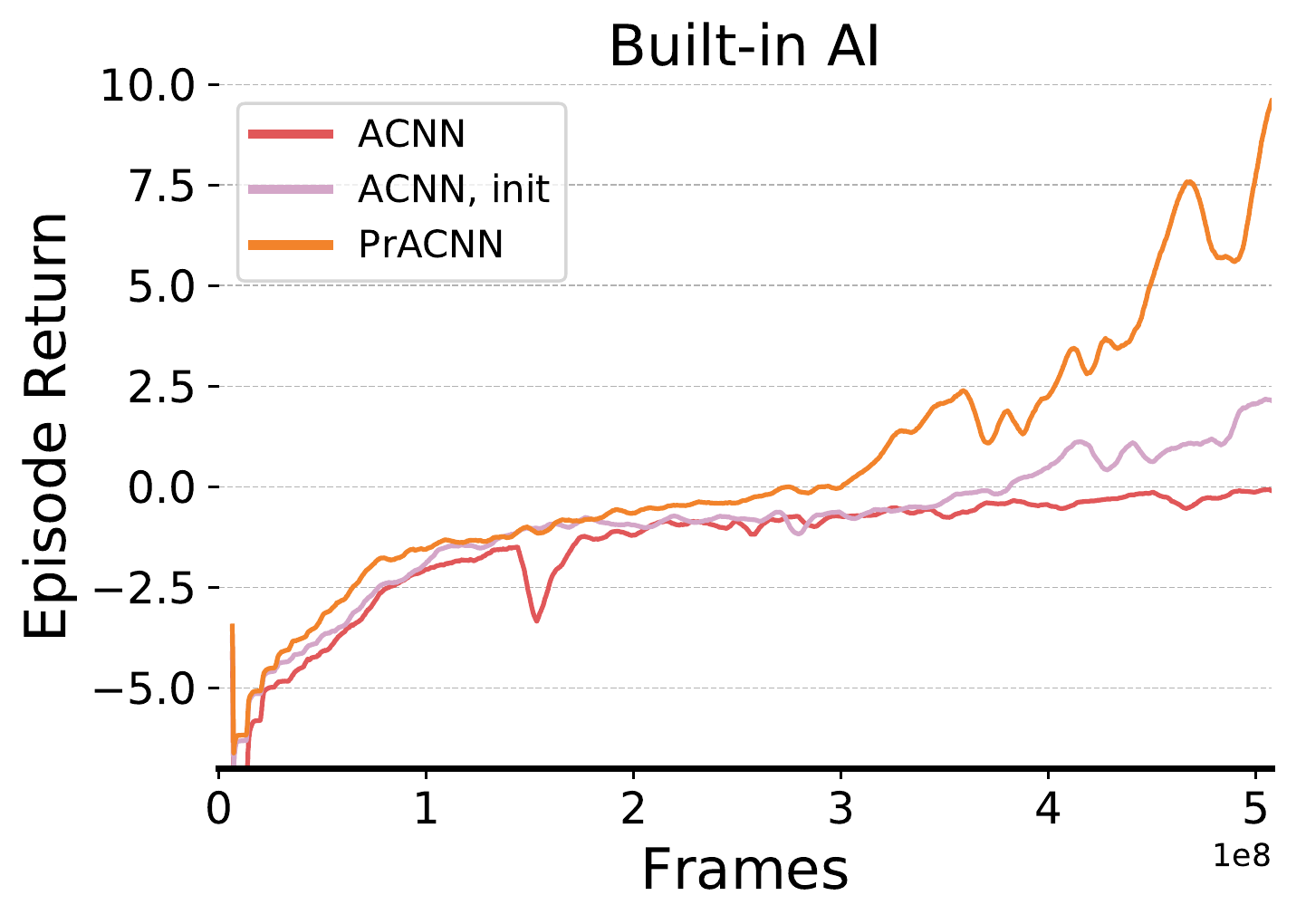}\hspace{0.35in}
\includegraphics[width=0.28\textwidth]{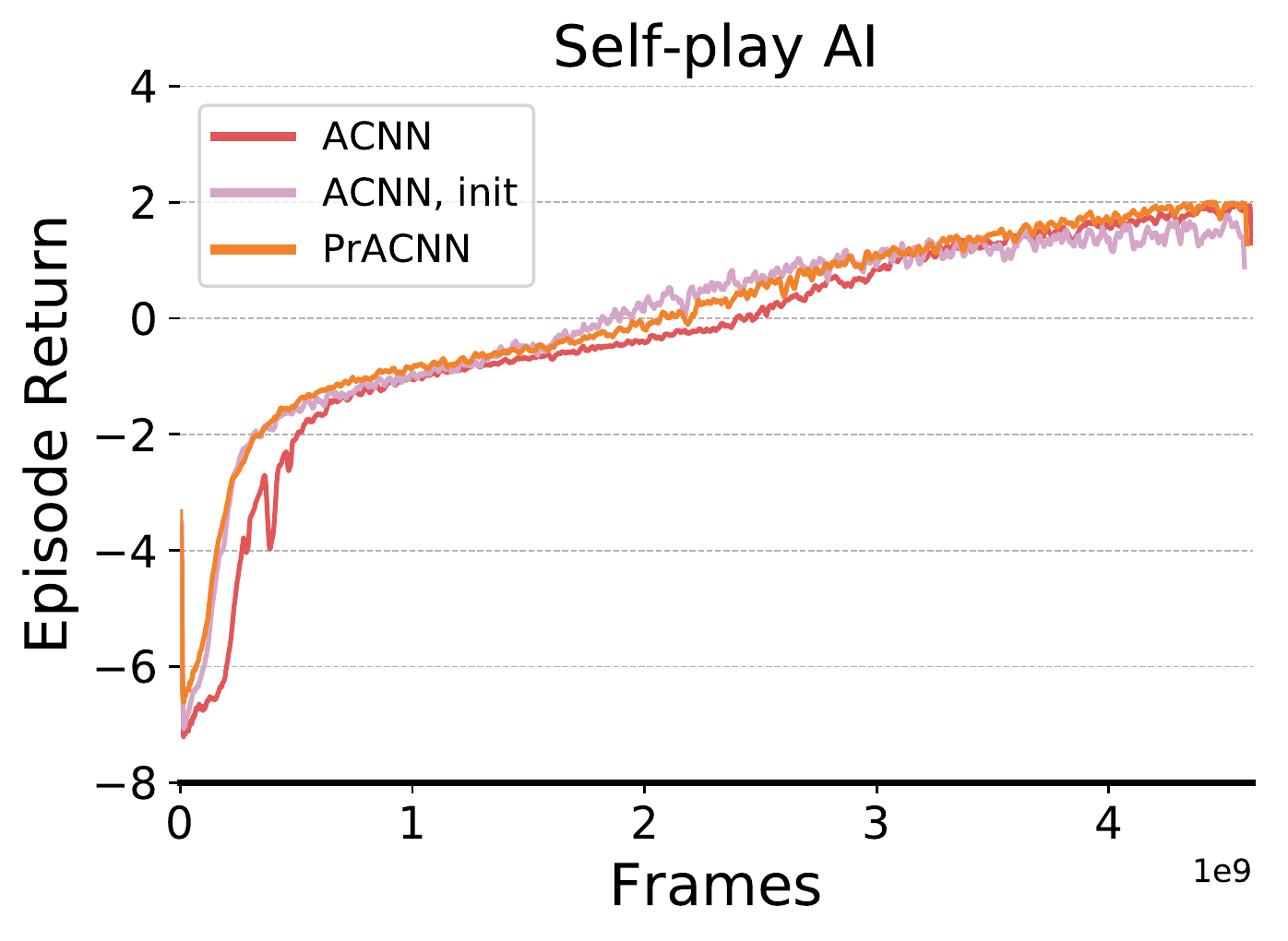}\hspace{0.35in}
\includegraphics[width=0.28\textwidth]{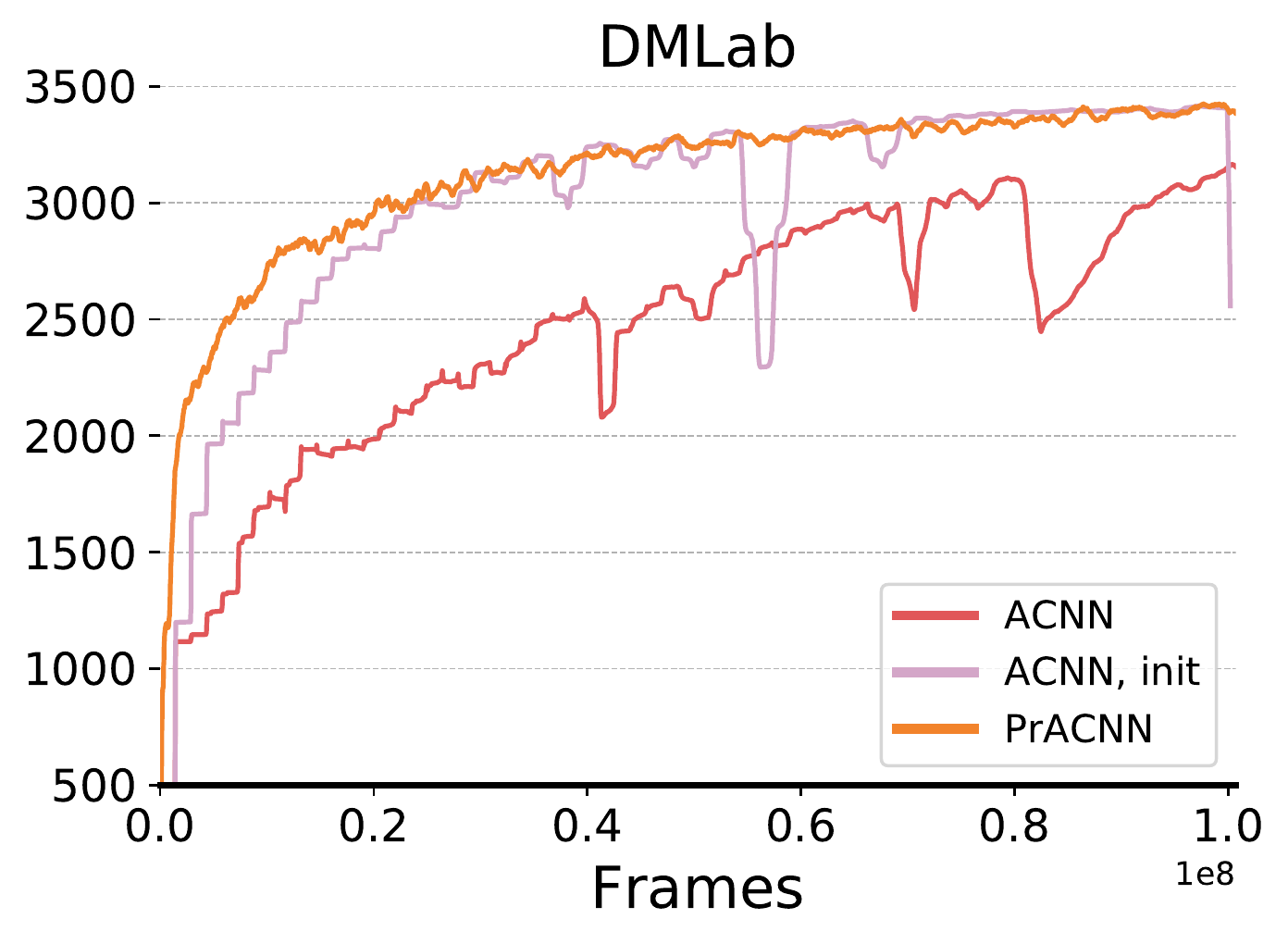}
}}\\
\subfloat[Comparison of {\CNN}, {\CNN} using initialization, and {\PrCNN} when trained with auxiliary loss.]{\label{fig:init3}{
\includegraphics[width=0.28\textwidth]{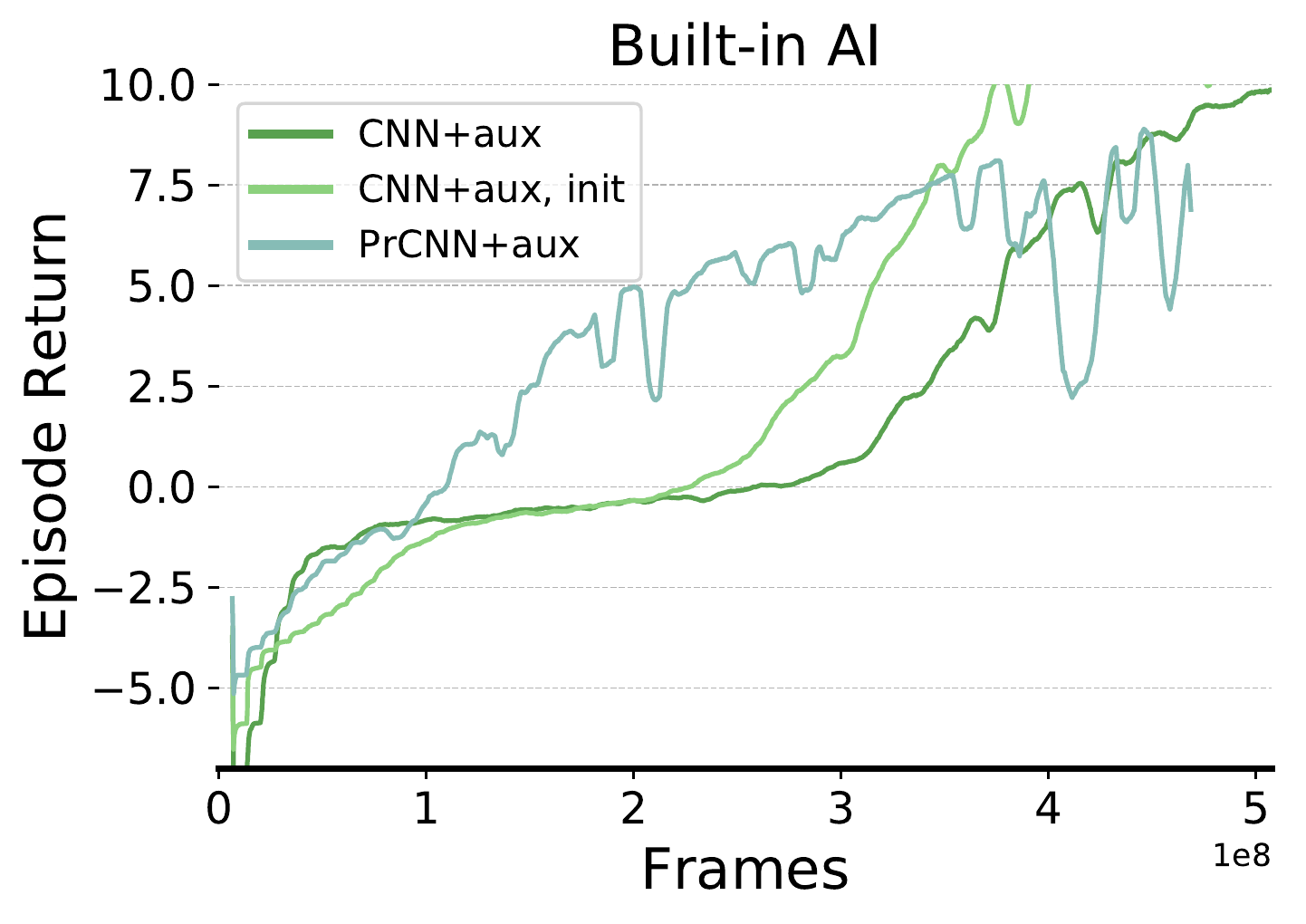}\hspace{0.35in}
\includegraphics[width=0.28\textwidth]{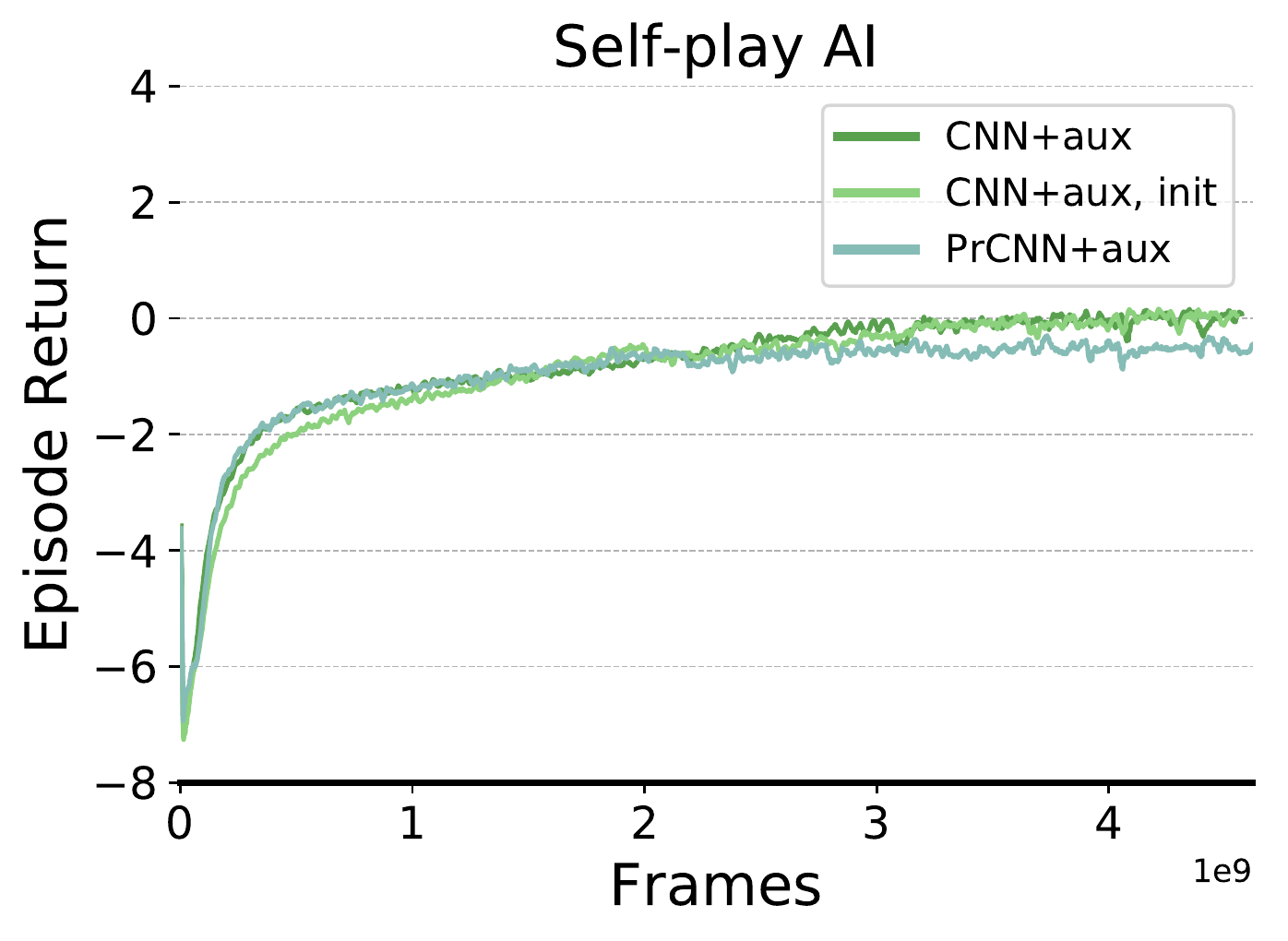}\hspace{0.35in}
\includegraphics[width=0.28\textwidth]{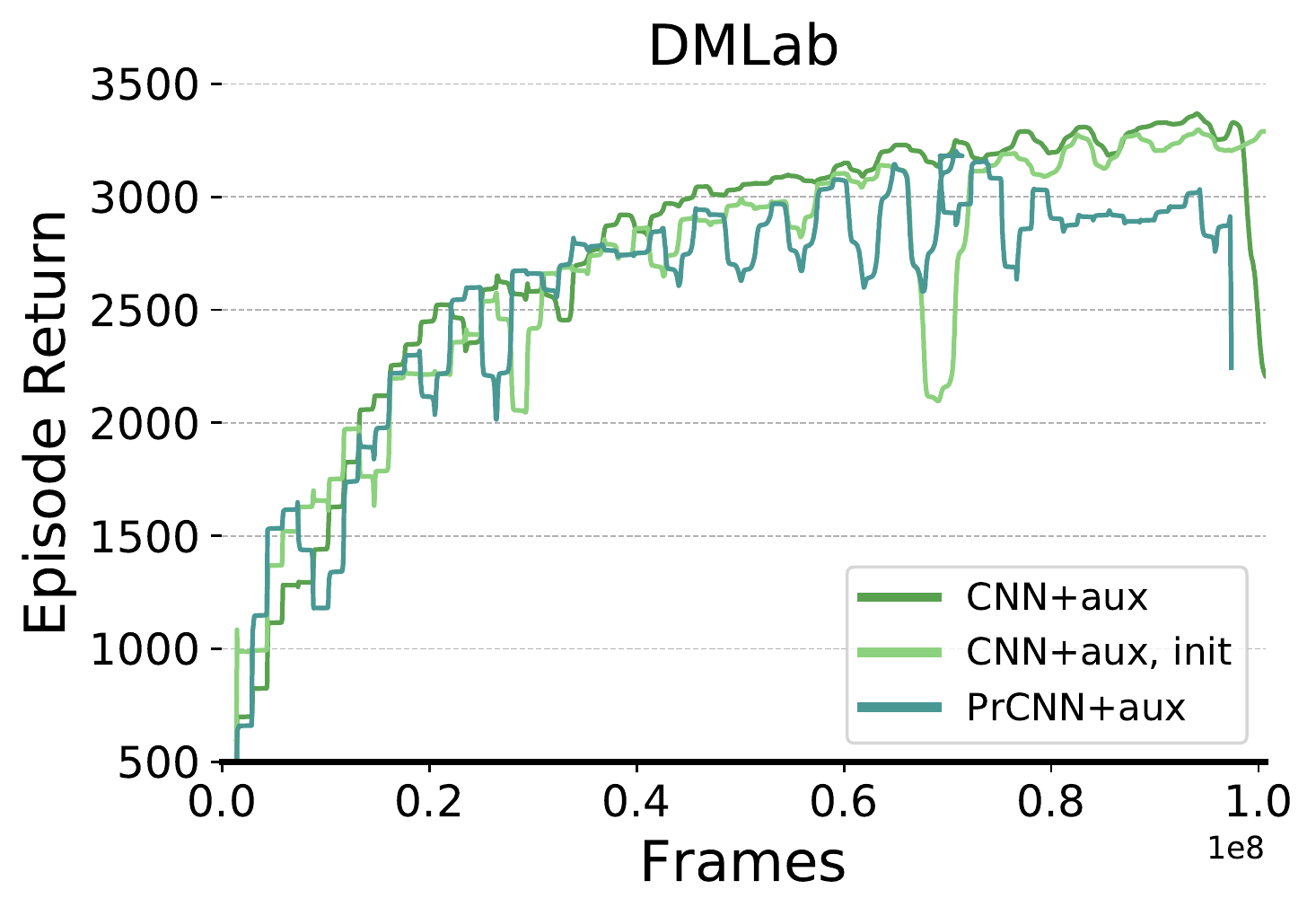}}}\\
\vspace{-0.1in}
\subfloat[Comparison of {\ACNN}, {\ACNN} using initialization, and {\PrACNN} when trained with auxiliary loss.]{\label{fig:init4}{
\includegraphics[width=0.28\textwidth]{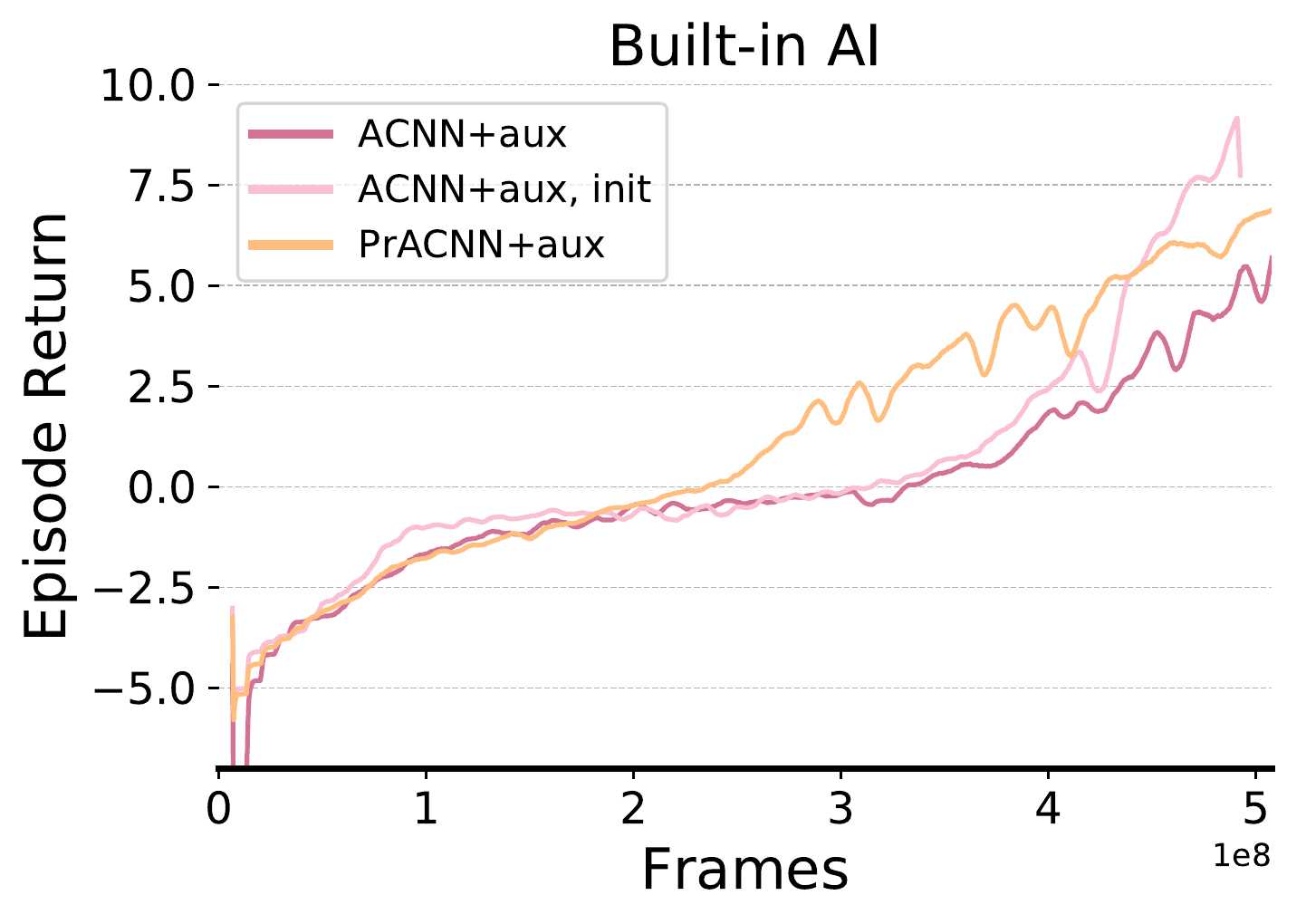}\hspace{0.35in}
\includegraphics[width=0.28\textwidth]{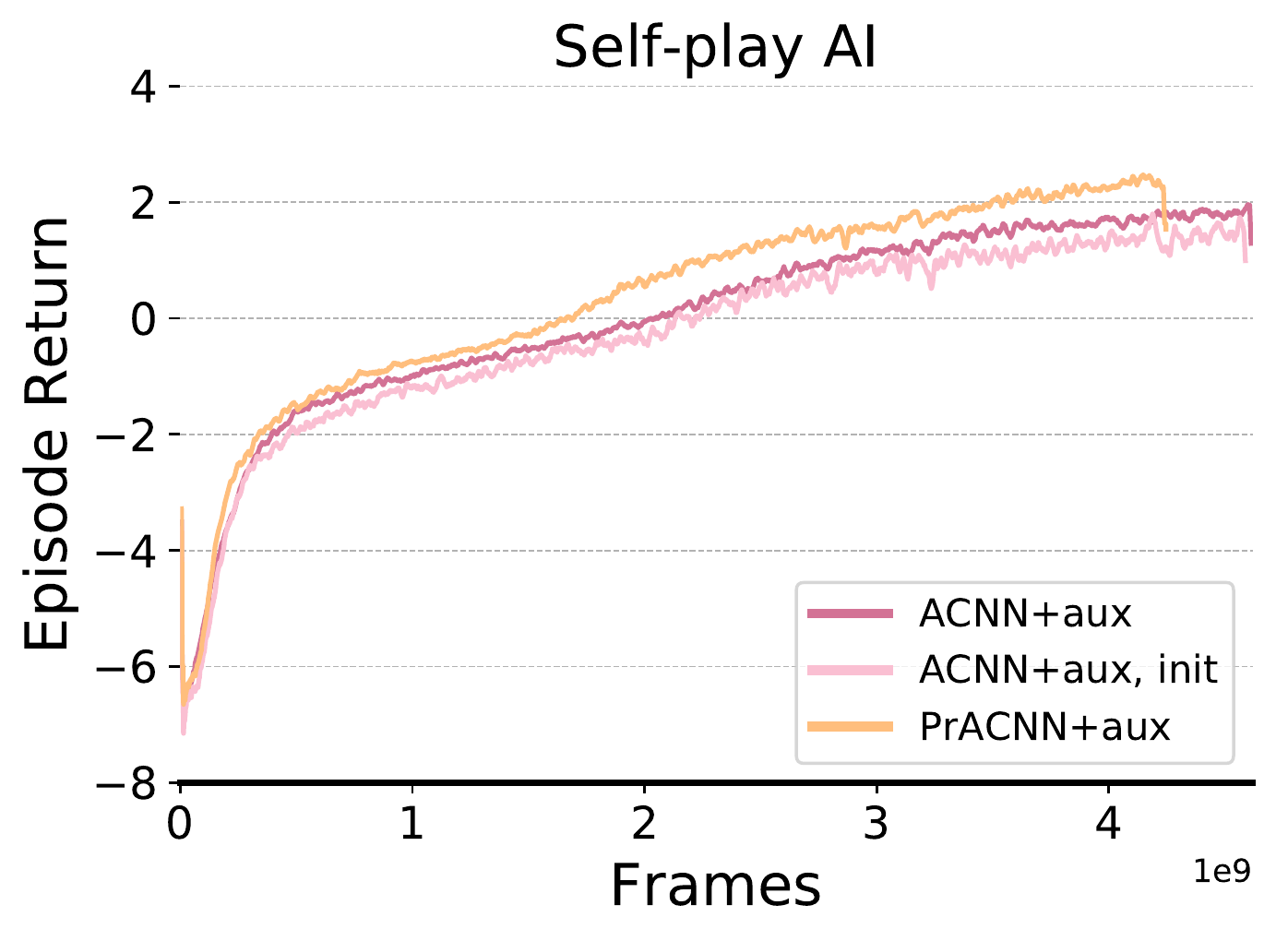}\hspace{0.35in}
\includegraphics[width=0.28\textwidth]{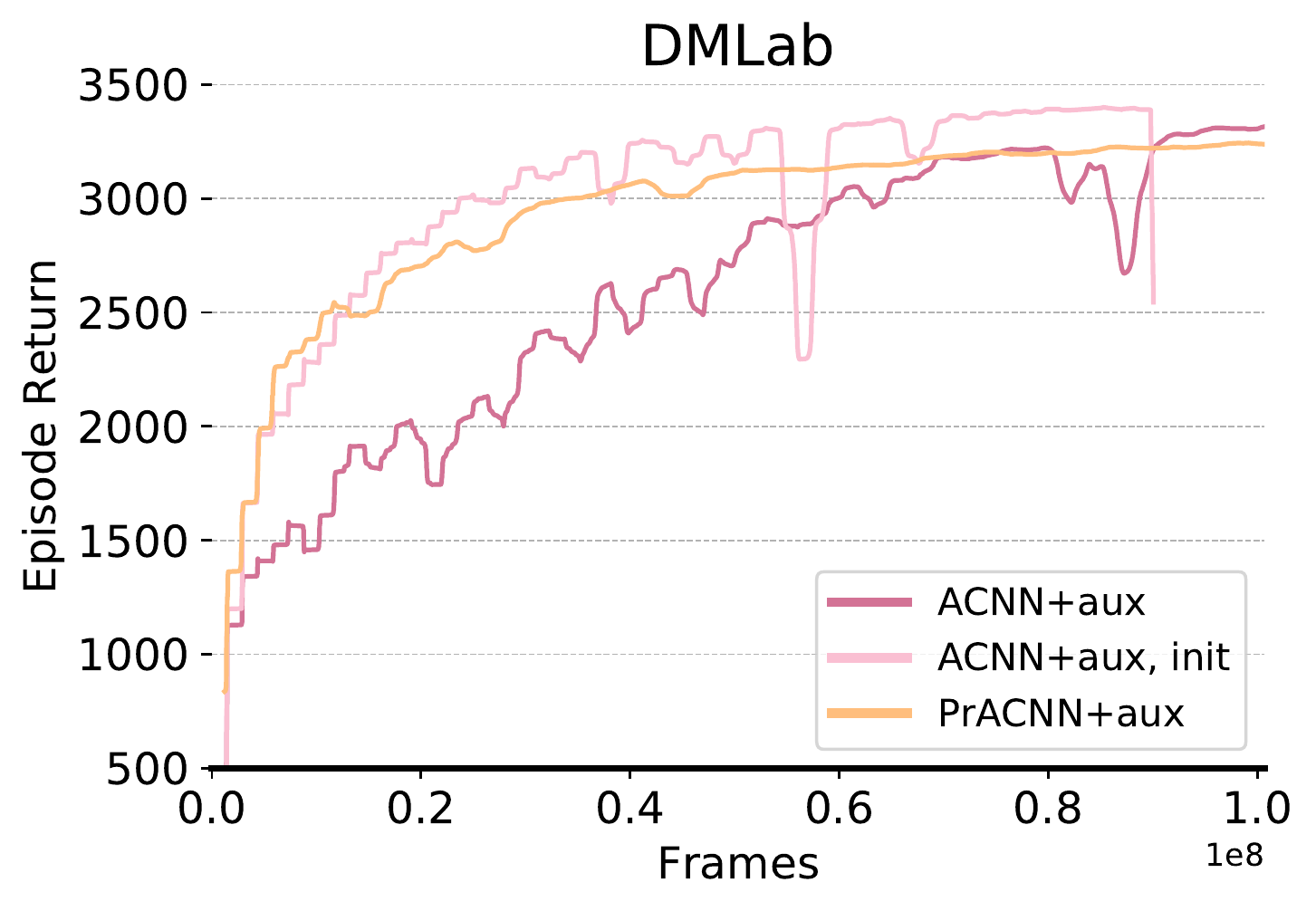}
}}\\
\caption{Comparison of training from scratch, using pre-trained models as initialization and as a frozen column in progressive networks, on top of (a) {\CNN}, (b) {\ACNN} models as well as (c) {\CNN} and (d) {\ACNN} with auxiliary loss. 
For (a) and (b), initialization and progressive networks show better or comparable performance as well as sample efficiency, with more evident effects on the simpler tasks, built-in AI and DMLab2D. The best performing model for 2 out of 3 tasks are {\PrACNN}.
The same trend is observed with auxiliary loss (c) and (d) but less evident, as the auxiliary loss and the pre-training loss likely carry overlapping information. 
%
Plots are averaged over 3 random seeds.
}
\label{fig:init}
\end{figure*}
\subsection{Representation Learning from Observations}\label{sec:observationlearning}
We collect replays without action labels for the \emph{unsupervised} representation learning from observations. 
For GRF built-in AI and DMLab2D ``Cleanup'', we record replays evenly over the course of the baseline RL training, from the early stage of training to convergence. 
For GRF self-play, we record checkpoints evenly throughout self-play training (details in Appendix 3) and sample checkpoints to play against each other. 
%
%
%
We collect approximately 300K frames from each task.
%
In principle, one can utilize \emph{any} reasonable replays. Unlike learning from demonstration, the unsupervised approaches we study do not require action annotations, enabling potential usage of observations for which it would be infeasible to label actions~\citep{schmeckpeper2019learning}. 

We learn two predictive models from observations as described in Section~\ref{sec:observation}, based on {\CNN} and {\ACNN}. The predictive objectives are optimized via negative log-likelihood (NLL) minimization using Adam~\citep{kingma2014adam} with default parameters in TensorFlow~\citep{abadi2016tensorflow} and a sweep over learning rate. 

%
The batch size is 32, and we train till convergence on the validation set. 
More training details and NLL results are summarized in Appendix 4. 
The attention module does not offer significant advantage in terms of NLL. 
We conjecture that because the location of each agent is predicted independently, the {\CNN} architecture has sufficient model complexity to handle the tasks used in our experiments.
\subsection{MARL and Main Results}\label{sec:mainexp}
%
The RL training procedures closely follow SEED RL~\citep{espeholt2019seed}. When adding the auxiliary loss, we sweep its weighting coefficient over a range of values as illustrated in Section~\ref{sec:implement}. Figures~\ref{fig:aux} and~\ref{fig:init} show plots of our core results, where the horizontal axis is the number of frames and the vertical axis the episode return. We train on 500M frames for GRF built-in AI, 4.5G for self-play AI and 100M for DMLab2D. All plots are averaged over 3 random seeds of the best performing set of hyperparameters. The rest of this section examines and discusses the experiments. 
\subsubsection{Essential Roles of Attention}
Attention is empirically shown to be essential to form complex strategies among agents. Figure~\ref{fig:aux} compares the baseline {\CNN} with the attention based {\ACNN}, all trained from random initialization. Figure~\ref{fig:init} compares {\CNN} and {\ACNN} with pre-training. For \emph{2 out of 3} environments, namely GRF self-play AI and DMLab2D ``Cleanup'', {\PrACNN} are the best performing models. 

For DMLab2D, {\PrACNN} converges significantly faster and more stably than its CNN counterpart without the attention module. 

For GRF, note that the most critical phase here is when the agent transitions from passively defending to actively scoring, i.e. when the scores cross the 0 mark. This transition signifies the agent's understanding of the opponent's offensive as well as defensive policy. After the transition, the agent may excessively exploit the opponent’s weaknesses. By examining the replays against the built-in AI and self-play AI (attached in the Supplementary Materials), we find it is possible to exploit the built-in AI with very simple tactics by tricking it into an offside position. Thus for built-in AI, CNN models' scoring more than ACNN in absolute term merely means the former is good at exploiting this weakness. On the other hand, when against self-play AI, ACNN models can pass the 0 mark to start winning, while none of the CNN models are able to achieve so. This strongly indicates that the attention module plays an essential role in providing the reasoning capacity to form complex cooperative strategies. 
%

Moreover, Figure~\ref{fig:visualattention} takes two players from {\ACNN} trained against the self-play AI and visualize their attention patterns. Green dots are the active player agents, and yellow ones are the other agents on the home team. We visualize the attention weights from one of the two heads. The intensity of the red circles surrounding the home agents reflect the weights. The most watched area is in the vicinity of the ball. E.g. in the 5th frame, one agent (top row) focuses on the player in possession of the ball, whereas the other agent (bottom row) is looking at the player to whom it is passing the ball. Full replays are in the Supplementary Materials.
\subsubsection{Effects of Agent-Centric Prediction}
\paragraph{The agent-centric auxiliary loss complements MARL.} Figure~\ref{fig:aux} compares the effects of adding the agent-centric auxiliary loss as described in Section~\ref{sec:aux} to the CNN and ACNN model, both trained from random initialization. As RL is sensitive to tuning, an incompatible auxiliary loss can hinder its training. 
%
In our experiments, however, the auxiliary loss for the most part improves the models' performance and efficiency, particularly for the CNN models. This suggests the agent-centric loss is supportive of the reward structure in the game. 

\paragraph{Unsupervised pre-training improves sample-efficiency.} 
We compare training the {\CNN} (Figure~\ref{fig:init1}) and {\ACNN} (Figure~\ref{fig:init2}) from scratch with the two ways of integrating unsupervised pre-training to MARL, namely weight initialization and progressive neural networks.
For a fair comparison, we use the same hyperparameter tuning protocol from baseline training to tune models involving pre-training. 
%
Both ways of integration provide significant improvements to sample-efficiency, especially on the simpler DMLab2D and GRF with the built-in AI. For some cases, the progressive models can achieve better performance and efficiency than those with the weight initialization. Even when the impact from pre-training is limited, it does not hurt the performance of MARL. 
Hence in practice it can be beneficial to perform a simple pre-training step with existing observation recordings to speed up downstream MARL. 
%
We repeat the same control experiments on RL models trained with the auxiliary loss in Figure~\ref{fig:init3} and Figure~\ref{fig:init4}. The same trend is observed, albeit to a lesser degree, which is as expected because the auxiliary objectives and pre-training carry overlapping information. 
%
%
%
%
%
\begin{figure*}[]
\centering
\includegraphics[width=0.24\textwidth]{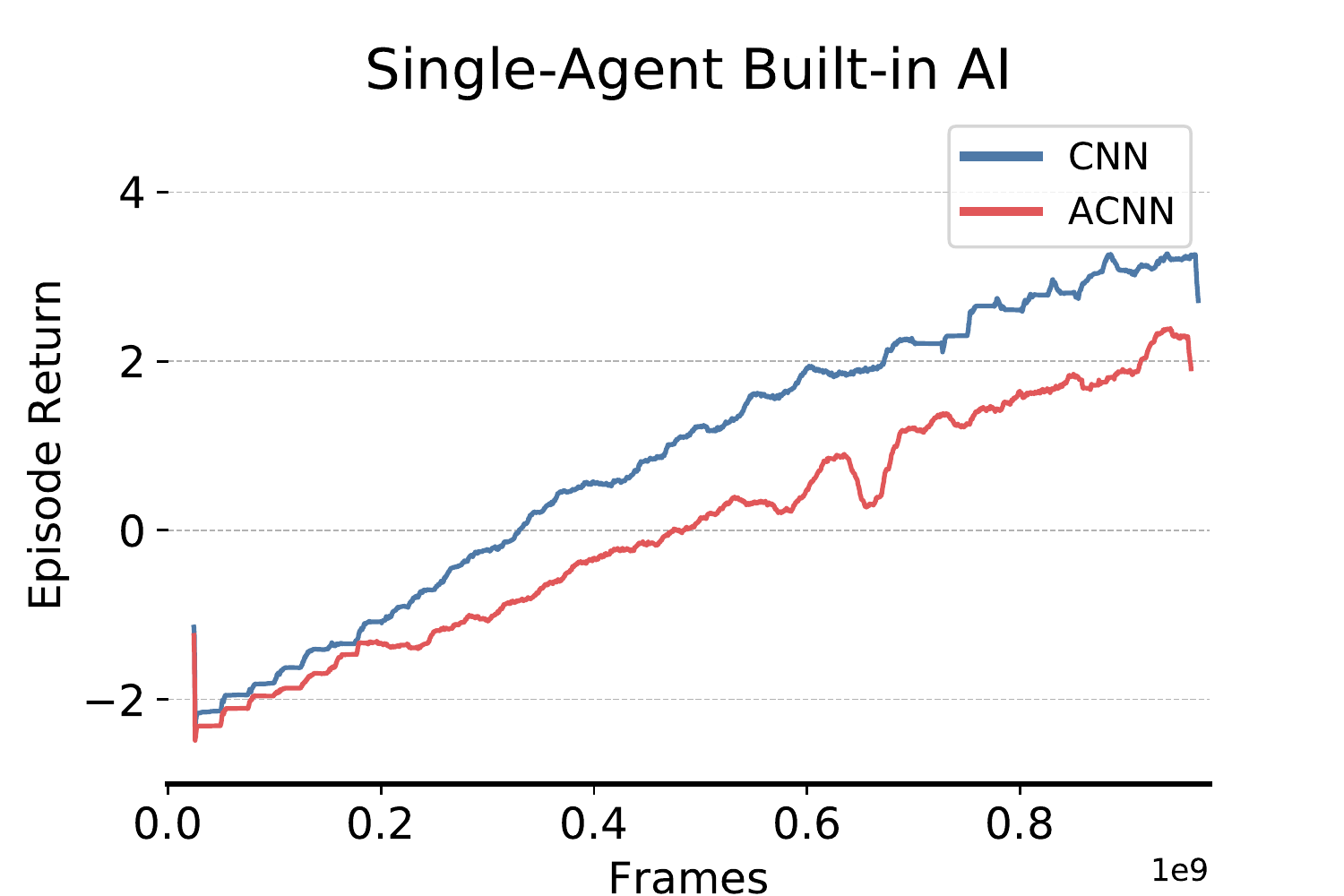}
\includegraphics[width=0.24\textwidth]{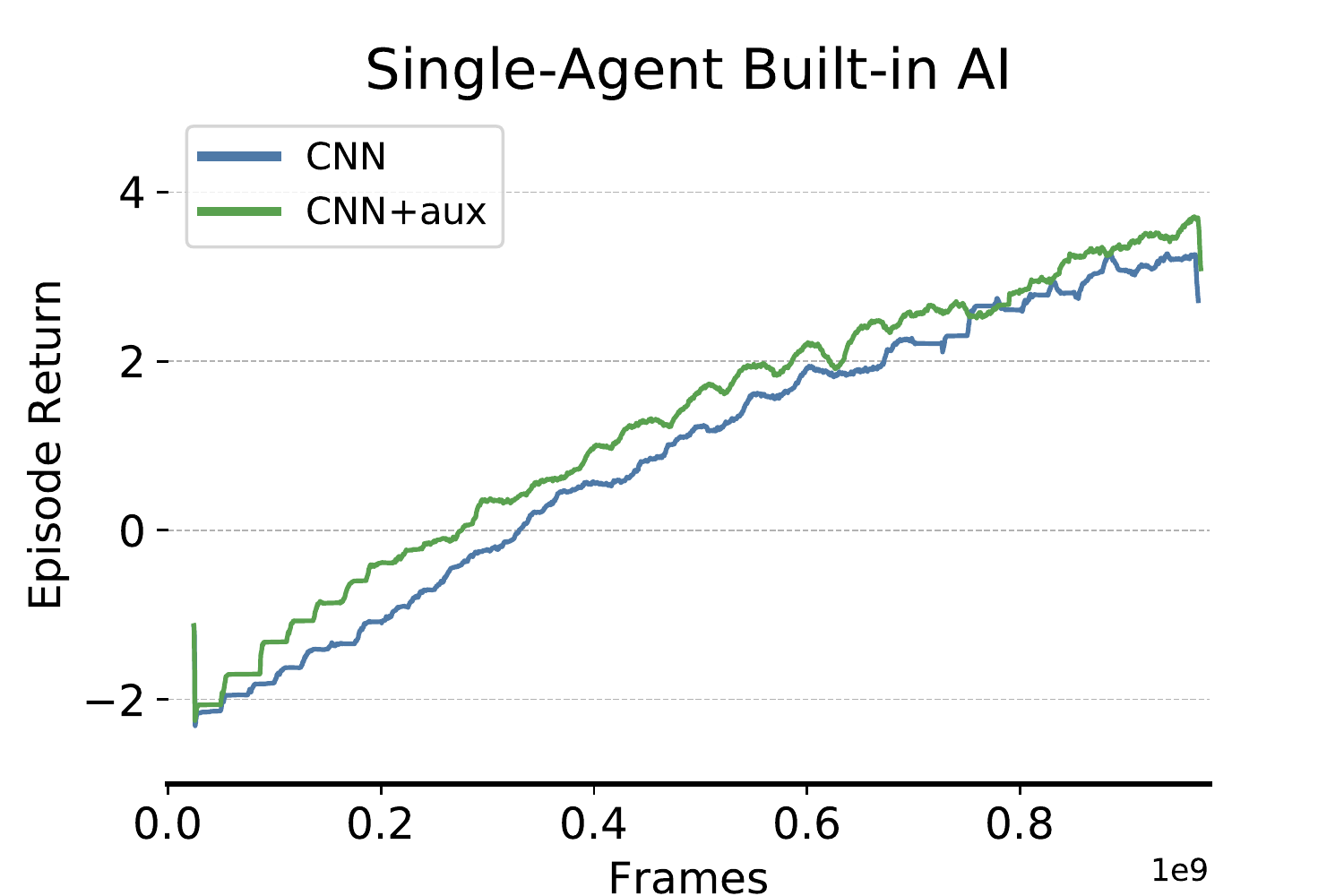}
\includegraphics[width=0.24\textwidth]{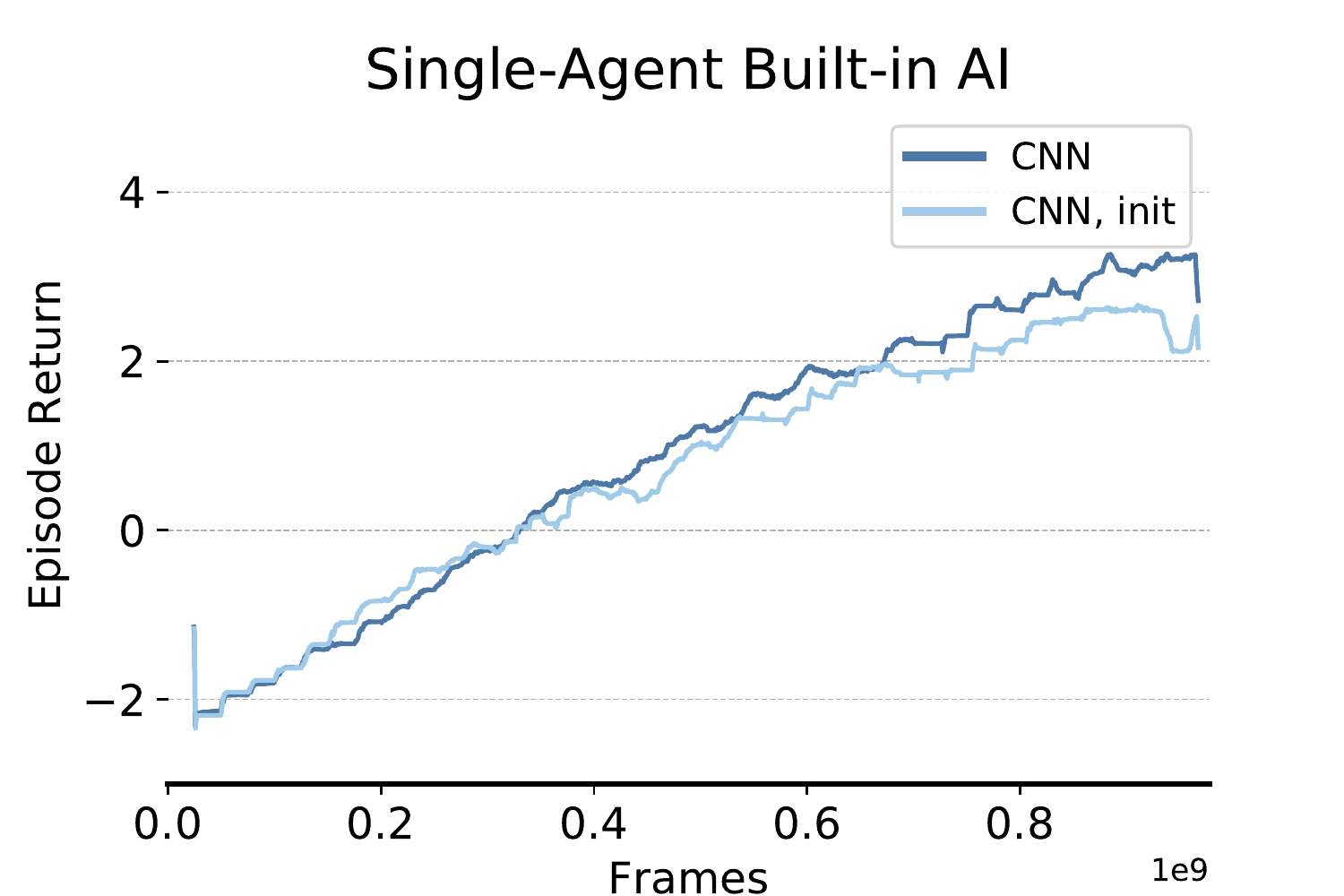}
\includegraphics[width=0.24\textwidth]{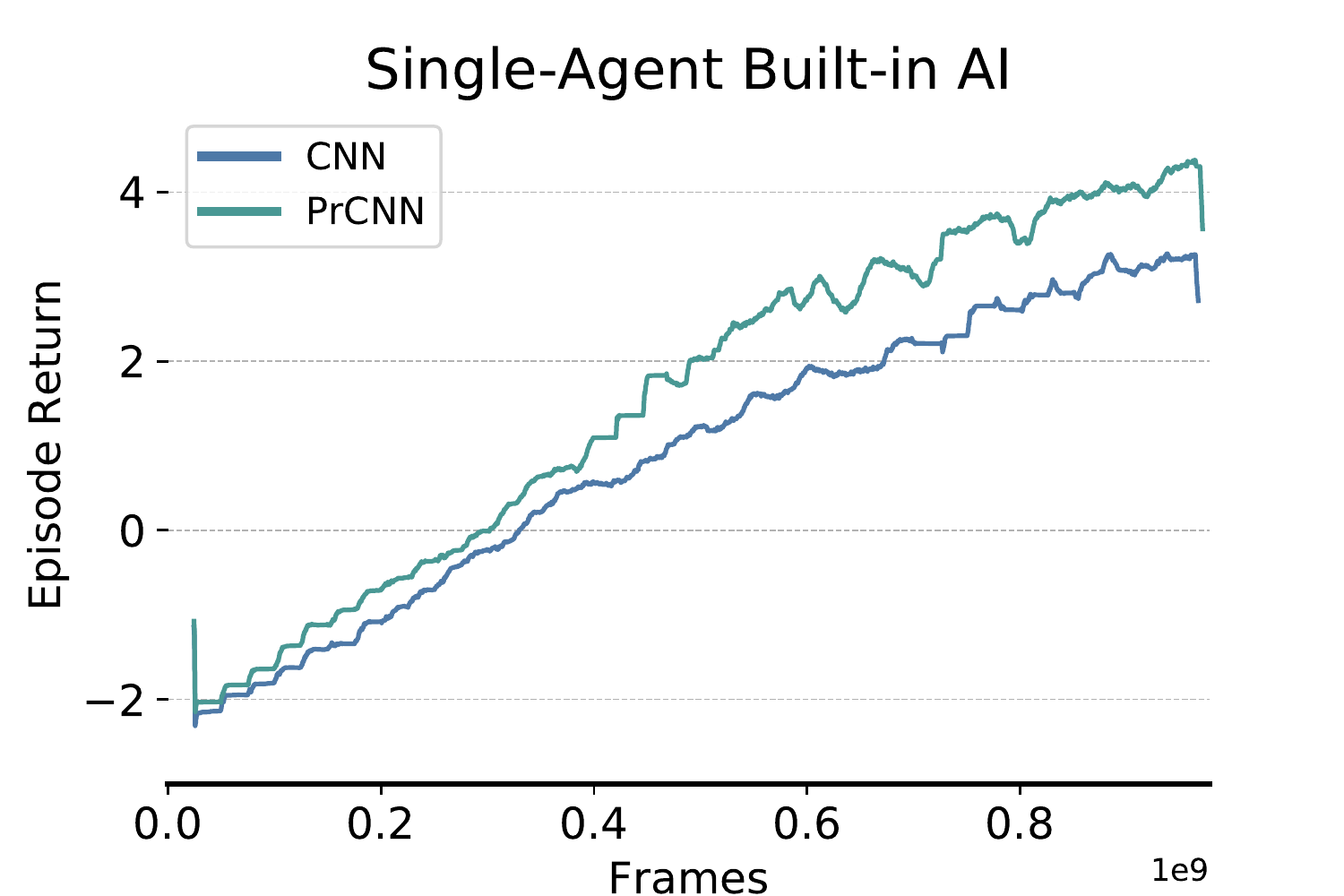}
\caption{Results of incorporating agent-centric inductive biases to \emph{single-agent RL} (GRF 11-vs-11 Hard Stochastic). The attention module is no longer optimal as cooperation matters less for single-agent RL. The auxiliary loss and pre-training from observations still help albeit to a lesser degree.}
\label{fig:single}
\end{figure*}
\begin{figure*}[]
\centering
\includegraphics[width=0.28\textwidth]{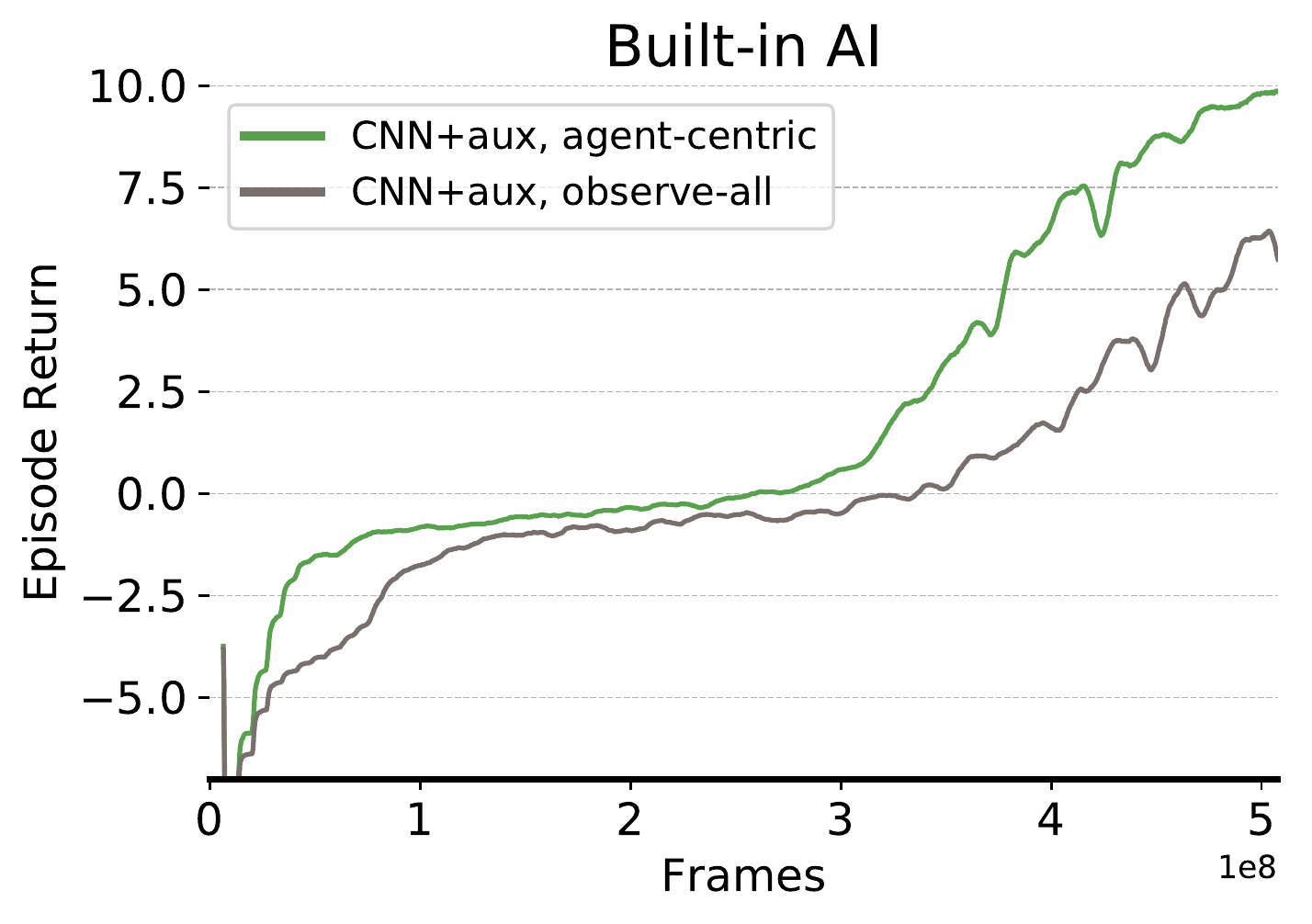}\hspace{0.35in}
\includegraphics[width=0.28\textwidth]{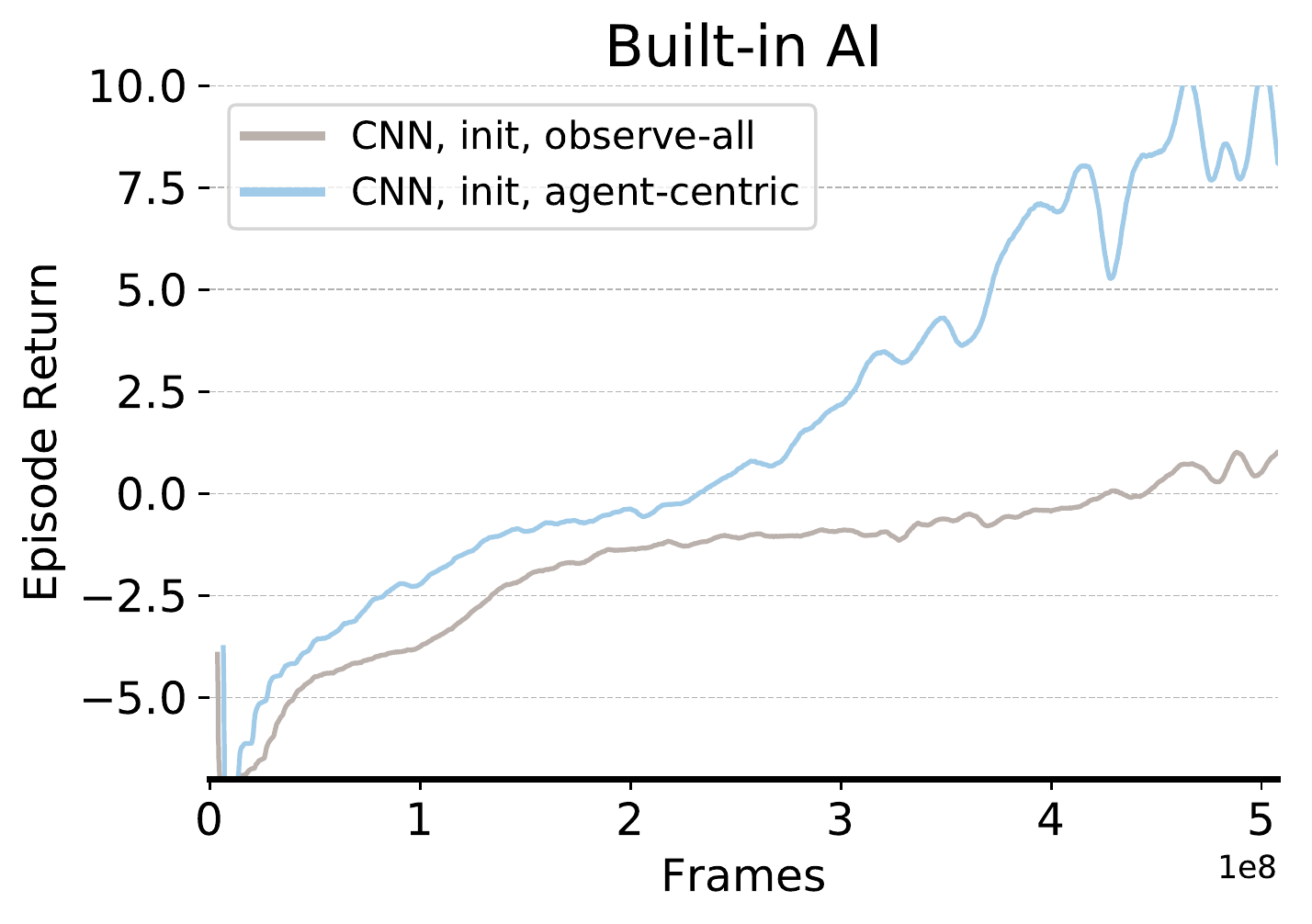}\hspace{0.35in}
\includegraphics[width=0.28\textwidth]{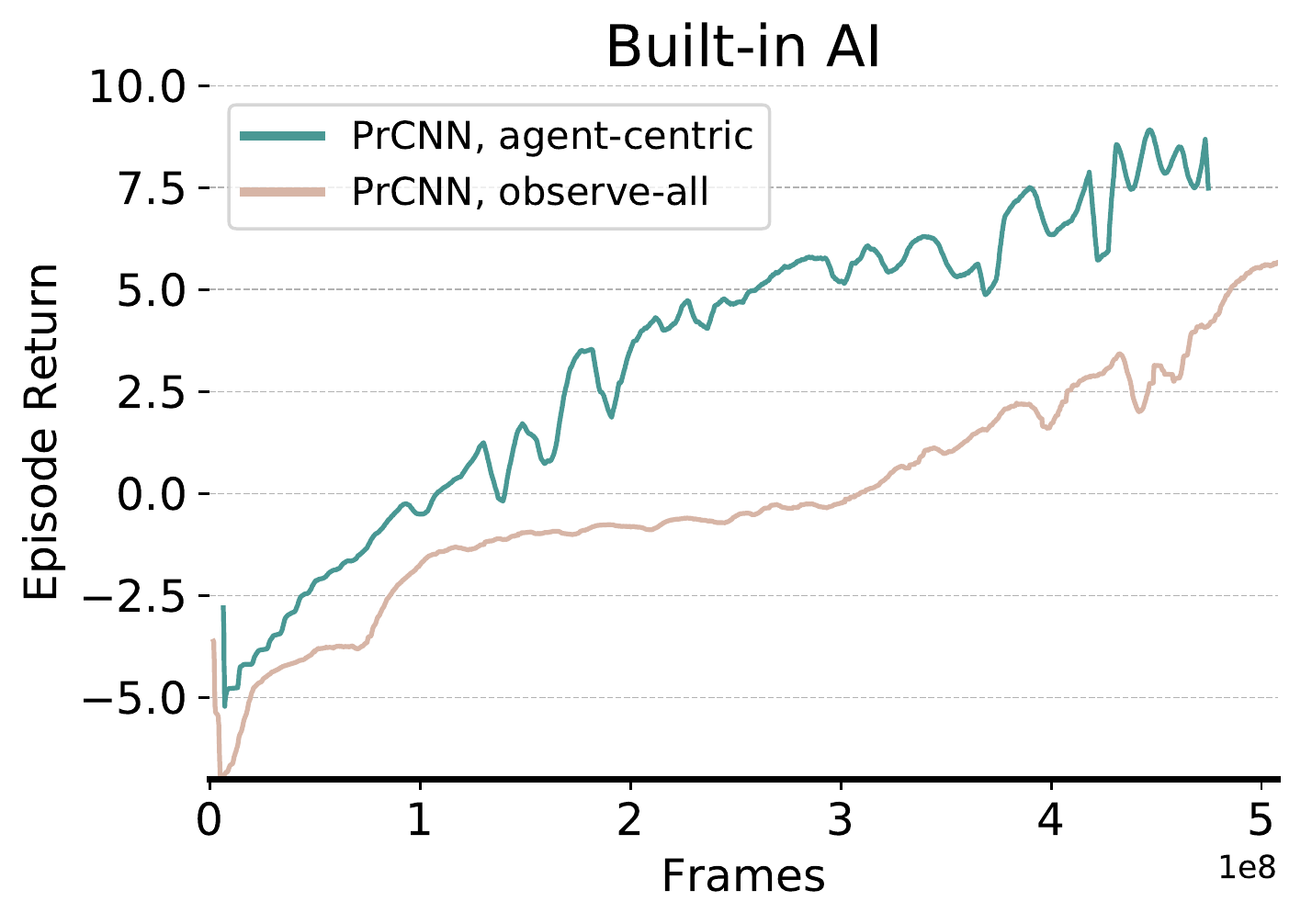}
\caption{Comparison between the agent-centric objective and the observe-all objective that predicts for all agents' location at once for MARL on GRF Built-in AI. The former exhibits clear advantage.}
\label{fig:observeall}
\end{figure*}
\subsection{Agent-Centric Representation Learning for Single-Agent RL}\label{sec:single}
We have demonstrated the efficacy of agent-centric representation learning for multi-agent RL. To evaluate whether similar conclusions holds for single-agent RL, we use the 11-vs-11 ``Hard Stochastic'' task~\citep{kurach2019google}, where only one player is controlled at a time (simiarly to FIFA).

Although the game logic has much in common between single and multiple players, the experimental outcome is different as shown in Figure~\ref{fig:single}. The attention module brings down baseline performance, likely because cooperation matters less here\textemdash the rest of the home players are controlled by the built-in AI\textemdash and the attention module is harder to optimize. The agent-centric prediction task still helps as auxiliary loss or for pre-training, but to a much limited extend. 
\subsection{Agent-agnostic Observe-All Representation Learning for MARL}\label{sec:observeall}
Finally, to verify the necessity for MARL representations to be \emph{agent-centric}, we implement an alternative agent-agnostic observation prediction objective, referred to as \emph{observe-all}, and test with playing against the GRF built-in AI. 
Concretely, the height and width predictive heads for observe-all output $72-$ and $96-$ dimension binary masks respectively, where 1 indicates player occupation and 0 vacancy, taking over \emph{all agents}. In this way, the prediction is agnostic of agent identity and can predict for all players at once.
First, we use the observe-all objective as an auxiliary loss to RL and train from scratch. Next we apply the observe-all objective to pre-training from observations, which is then used for MARL, either via weight initialization or as a progressive frozen column. Figure~\ref{fig:observeall} clearly shows the agent-centric auxiliary loss is more competitive than the observe-all one in all setups. It confirms that the agent-centric nature of the prediction task is indeed important for MARL.
\begin{table*}[t]
\begin{minipage}[b]{0.7\linewidth}\centering
\resizebox{0.99\linewidth}{!}{
\begin{tabular}{ccccc|cccc|c}
& {\CNN} & +aux. & +init. & +prgs. & \footnotesize{{\ACNN}} & +aux. & +init. & +prgs. &rating\\
{\CNN} & \backslashbox[7mm]{}{}& 5/8/7 & 9/1/10 & 7/5/8 & 4/4/12 & 5/2/13 & 2/7/11 & 3/5/12 & ${-}323$ \\
+aux. & 7/8/5 & \backslashbox[7mm]{}{} & 6/4/10 & 6/1/13 & 3/4/13 & 2/4/14 & 2/8/10 & 2/5/13 &${-}420$\\
+init. & 10/1/9 & 10/4/6 & \backslashbox[7mm]{}{} & 6/5/9 & 5/5/10 & 2/5/13 & 3/6/11 & 4/6/10 &${-}283$\\
+prgs. & 8/5/7 & 13/1/6 & 9/5/6 & \backslashbox[7mm]{}{} & 2/8/10 & 2/4/14 & 5/1/14 & 2/5/13 &${-}322$\\ \hline
\footnotesize{{\ACNN}} & 12/4/4 & 13/4/3 & 10/5/5 & 10/8/2 & \backslashbox[7mm]{}{} & 8/4/8 & 7/7/6 & 10/6/4 &$\mathbf{147}$\\
+aux. & 13/2/5 & 14/4/2 & 13/5/2 & 14/4/2 & 8/4/8 & \backslashbox[7mm]{}{} & 11/3/6 & 7/5/8 &$\mathbf{182}$\\
+init & 11/7/2 & 10/8/2 & 11/6/3 & 14/1/5 & 6/7/7 & 6/3/11 & \backslashbox[7mm]{}{} & 8/4/8 &$24$\\
+prgs. & 12/5/3 & 13/2/5 & 10/6/4 & 13/5/2 & 4/6/10& 8/5/7 & 8/4/8 & \backslashbox[7mm]{}{} &$35$\\
\end{tabular}}
\end{minipage}
\begin{minipage}[b]{0.28\linewidth}
\centering
\resizebox{0.99\linewidth}{!}{\begin{tabular}{cc|c}
 \multicolumn{3}{c}{GRF Leaderboard Results}  \\ \hline
Agent&Opponent&Rating\\
\hline
{\ACNN}+aux &CNN-v1/v2&$\mathbf{1992}$\\
{\ACNN}&CNN-v1/v2&$1841$\\
{\CNN}-v1& itself &$1659$\\
{\CNN}-v2& itself &$1630$\\
{\CNN}+aux&Built-in&$1048$\\
Built-in&NA&$1000$\\
\end{tabular}}
\end{minipage}
\caption{
\small{\textbf{Left:} selected agents play 20 matches among each other, all trained against the self-play AI. Each entry records win/tie/loss between row agent and column agent. Ratings are estimated ELO scores. {\ACNN}s shows clear superiority over {\CNN}s. {\ACNN} trained from scratch generalizes better. \textbf{Right:} GRF Multi-Agent public leaderboard results by the time of model submission. Opponent refers to the policy the agent trained against. Each submitted agent plays 300 games.  {\ACNN}s trained against self-play AI perform the best. The {\CNN} trained against built-in AI performs poorly and does not generalize. The self-play AI used in our experiments {\CNN}-v1/v2 are clearly superior to the built-in AI.} 
}\label{tab:tournament}
\end{table*}

\section{GRF Agent Tournament and Public Leaderboard}\label{sec:lb}
Table~\ref{tab:tournament} (Left) investigates how various agent-centric representation learning components generalize by hosting a tournament among selected agents. We include agents trained from scratch, from scratch plus auxiliary loss, from initialization plus auxiliary loss, and with progressive column plus auxiliary loss.
%
All are trained against self-play AI for a total of 4.5 billion frames and each play 20 matches against another. 
Each entry records win/tie/loss between row agent (home) and column agent (opponent).
We also estimate their ELO ratings (for details see Appendix 6). Clearly, {\ACNN} based models outperform {\CNN} models, corroborating the claim that the agent-centric attention module enhances generalization. Meanwhile, the {\ACNN} models using pre-training are inferior to the ones trained from scratch.
It suggests that although pre-training speeds up convergence, it can also limit the model's ability to generalize. 

Finally, we upload our best performing agent trained against the built-in AI, i.e. {\CNN} with auxiliary loss, and best agents against the self-play AI, i.e. {\ACNN} and {\ACNN} with auxiliary loss, to the public GRF Multi-agent League~\citep{leaderboard}. For comparison, we also upload the two self-play AI models used in our experiments, i.e. {\CNN}-v1 and {\CNN}-v2. Each of the submitted agents play 300 games against agents submitted by other participants and their ELO ratings are listed in Table~\ref{tab:tournament} (Right).
%
The self-play AI agents deliver a decent performance, showing clear advantages over the built-in AI. The agents trained against the self-play AI overall perform the best. In contrast, the agent trained against the built-in AI, although dominating the built-in AI, is extremely fragile against other agents. This supports our observation that the built-in AI can be exploited with little cooperation. It is also worth mentioning that, at the time of submission, our {\ACNN} agent with auxiliary loss ranks top 1 on the learderboard.

\vspace{-0.15in}
\section{Conclusions}
\vspace{-0.1in}
We propose to integrate novel agent-centric representation learning components, namely the agent-centric attention module and the agent-centric predictive objective, to multi-agent RL. In experiments, we show that the attention module leads to complex cooperative strategies and better generalization. In addition, leveraging the agent-centric predictive objectives as an auxiliary loss and/or for unsupervised pre-training from observations improves sample efficiency. 
\section*{Acknowledgements}
We would like to thank Charles Beattie for help on DMLab2D and Piotr Stanczyk on GRF. We would also like to thank Thomas Kipf, Alexey Dosovitskiy, Dennis Lee, and Aleksandra Faust for insightful discussions.
\bibliography{reference}
\end{document}